\documentclass[runningheads]{llncs}

\usepackage{eccv}

\usepackage{eccvabbrv}

\usepackage{graphicx}
\usepackage{booktabs}
\usepackage[accsupp]{axessibility}  % Improves PDF readability for those with disabilities.

\usepackage[breakable]{tcolorbox}
\usepackage{multirow}
\usepackage{array}
\usepackage{tabularx}
\usepackage{hyperref}

\usepackage{orcidlink}

\begin{document}

\title{EM3M: An Electron Micrograph Dataset for Microstructural Segmentation and Generation} 

\titlerunning{EM3M}

%\author{Nan Wang \and Zhiyi Xia \and Yiming Li \and Shi Tang \and Zoe Fan \and  Xi Fang \and Haoyi Tao \and Siyuan Zhang \and Guolin Ke \and Yanhui Hong}

\author{Nan Wang\thanks{Equal contribution.} \and 
Zhiyi Xia\textsuperscript{*} \and 
Yiming Li\textsuperscript{*} \and 
Siyuan Zhang \and 
Shi Tang \and 
\\ Zoe Fan \and  
Xi Fang \and 
Haoyi Tao \and 
Guolin Ke \and 
Yanhui Hong\thanks{Corresponding author.}}

\authorrunning{N.~Wang et al.}

\institute{DP Technology, Beijing, China\\
\email{\{wangnan01, hongyanhui\}@dp.tech}}

\maketitle

\begin{abstract}
Quantitative microstructural characterization is fundamental to materials science, and electron micrographs (EMs) provide indispensable high-resolution insights. However, progress in deep learning-based analysis of EMs has been hampered by the scarcity of large-scale, expert-annotated public datasets. To address this issue, we introduce EM3M, a large-scale and multimodal dataset for instance-level understanding of EMs.
EM3M comprises 5,091 high-quality EMs, approximately \textbf{3 million} instance segmentation annotations, and image-level textual descriptions with disentangled attributes. The dataset is constructed through a rigorous multi-stage curation and validation pipeline, with comprehensive statistical analyses to ensure reliability and reproducibility.
Building upon these curated image-text pairs, we further provide a text-to-image diffusion model that serves as a controllable data augmentation engine, demonstrating that synthetic augmentation consistently improves downstream  segmentation performance. 
To establish a systematic benchmark, we evaluate representative instance segmentation methods on EM3M. Our results reveal that conventional detection-based and query-based methods struggle with the extreme instance densities and textural complexities inherent in EMs. We additionally provide an optimized flow-based baseline to facilitate fair comparison and future research.
%To establish a rigorous benchmark, we evaluate representative instance segmentation methods and provide a strong flow-based baseline (UniEM-Net). Quantitative and qualitative experiments demonstrate that conventional detection-based and query-based methods struggle with the extreme instance densities and textural complexities inherent in EMs, whereas our optimized baseline sets a high standard for future research.
%adapted to handle the extreme instance density and scale variations inherent in EMs. Quantitative experiments  show that while conventional methods struggle, our optimized baseline sets a high standard for future research.
%EM3M, together with the evaluation toolkit, pretrained baselines, and the generative engine, will be publicly available to accelerate progress in automated materials analysis.
EM3M\footnote{Dataset: \url{https://huggingface.co/datasets/UniParser/EM3M}}, the generative engine\footnote{Generation: \url{https://huggingface.co/UniParser/EM3M-Gen}}, and an online demo\footnote{Segmentation demo: \url{https://www.bohrium.com/apps/uni-aims}} are publicly available to support future research in automated materials analysis.

\end{abstract}    
\section{Introduction}
\label{sec:intro}
Quantitative microstructural characterization underpins advances in materials design, diagnostics, and fabrication. Electron microscopy techniques such as scanning electron microscopy (SEM) and transmission electron microscopy (TEM) are widely employed to capture microstructural details, enabling precise measurement of grain size, density, and distribution~\cite{brandon2013microstructural,callister2022fundamentals,goldstein2017scanning}. However, large-scale EM analysis remains challenging due to the need for domain expertise, costly licensing, and limited transferability of existing tools, such as ImageJ~\cite{collins2007imagej} and MountainsSEM.

%Although deep learning has introduced a new paradigm for automated EM analysis~\cite{choudhary2022recent}, 
Despite the promise of deep learning for automated EM analysis~\cite{choudhary2022recent},
most models are typically developed for specific tasks and struggle to generalize across the vast heterogeneity of real-world EM data. Previous studies have applied learning-based methods to nanomaterial classification~\cite{aversa2018first}, microstructural segmentation~\cite{okunev2020nanoparticle, emps, stuckner2022microstructure, shi2022automatic, bals2023deep, wang2021dynamic}, and micrograph generation~\cite{vagenknecht2023deep,bals2023artificial,ruhle2021workflow}. However, these efforts are often confined to particular imaging modalities, operational modes, or material types, and therefore lack the universality required to address the broader spectrum of EMs. Additionally, costly EM acquisition, expert annotation, and IP/privacy restrictions~\cite{shi2022u,ma2021data,opatz2025sharing} limit the availability of large, diverse public datasets, which in turn hinders the development of Segment Anything-style foundation models (SAM)~\cite{kirillov2023segment} for materials science.

The multimodal domain presents additional challenges. A comprehensive understanding of material microstructures requires correlating visual patterns with precise scientific interpretations—yet existing datasets fall short. For instance, MMSci~\cite{li2024mmsci} has harvested vast volumes of micrographs and associated text from scientific literature, but its figures are post‑processed for publication and its captions are often noisy and incomplete. 
Datasets such as MicroVQA~\cite{burgess2025microvqa} and MicroBench~\cite{lozano2024micro} include electron microscopy modalities but focus on biological and medical domains, leaving a gap for materials-oriented multimodal resources.

\begin{figure*}[tb]
    \centering
    \includegraphics[width=0.99\textwidth]{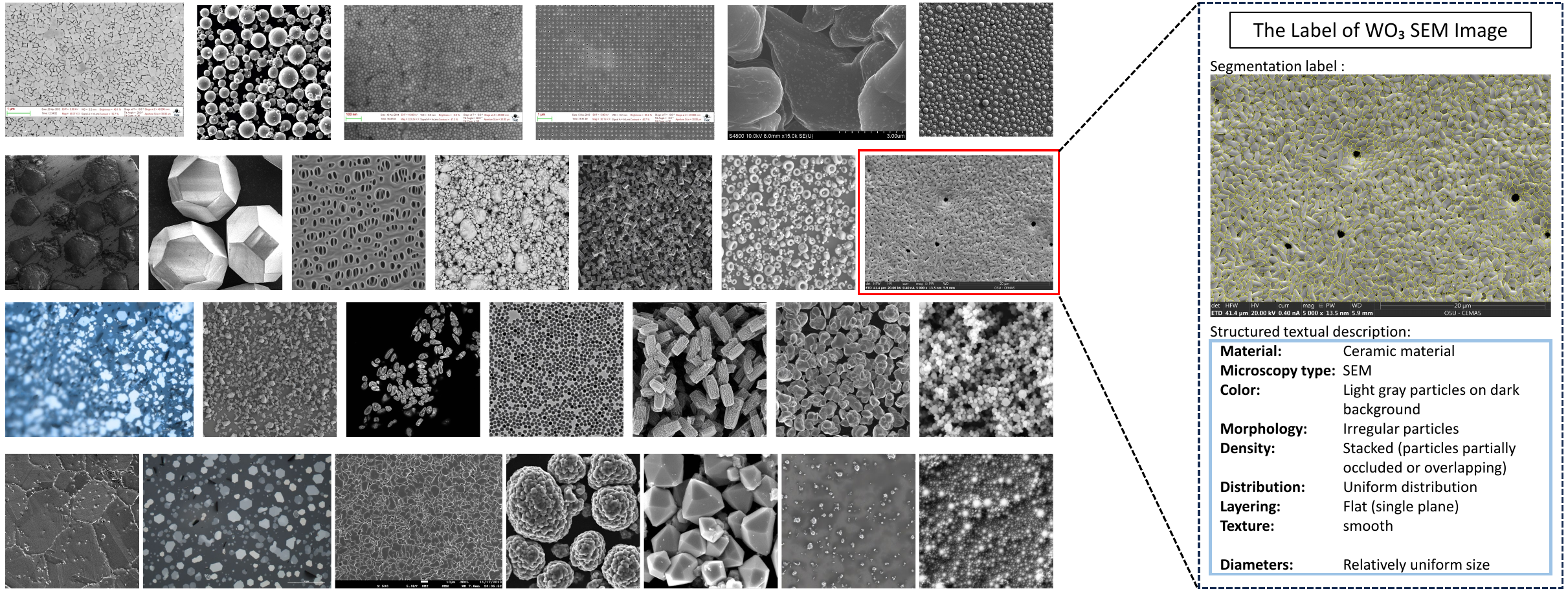} % Reduce the figure size so that it is slightly narrower than the column. Don't use precise values for figure width.This setup will avoid overfull boxes.
    \caption{Visualization of the EM3M. \textbf{Left}: Diverse raw EMs from different materials and morphologies. \textbf{Right}: A WO$_3$ sample exhibiting dense particulate features, along with its instance segmentation masks containing \textbf{2,445} annotated particles, and structured textual annotations describing material type, morphology, diameter distribution, and other attributes.}
    \label{fig:sample_image}
\end{figure*}

To address these issues, we present EM3M, a large-scale and multimodal EM dataset designed for instance-level materials understanding. 
Through extensive academic and industrial collaborations, we curated a collection of 5,091 high-quality electron micrographs, covering a wide range of material types, fabrication processes, and imaging modalities across SEM and TEM instruments, which capture representative microstructural morphologies.
Crucially, via a dedicated human-in-the-loop pipeline, we have annotated approximately \textbf{3 million} instance masks in total, achieving unprecedented annotation density with certain images exceeding 2,000 instances. Fig.~\ref{fig:sample_image} illustrates the diversity of our dataset and the richness of its annotations. Unlike MMSci, EM3M is built on raw experimental micrographs and uses an expert-defined protocol to produce structured, disentangled attribute descriptions for each image.
To further enhance the dataset's utility, we leverage the curated image-text pairs to train a text-to-image diffusion model that functions as a controllable augmentation engine, enabling scalable synthesis of realistic micrographs for the annotation pipeline. We also note that the decoupled attributes provide the secondary benefit of creating novel samples through recombination. While our primary focus is dataset construction, the structured descriptions provide a natural foundation for future materials-oriented vision–language modeling.

\begin{table}[tb]
\caption{
Summary of EM datasets. *Estimated upper bound of EMs in MMSci. \textsuperscript{†}Images were extracted from literature figures rather than raw experimental data.
Abbreviations: Mixed–synthetic and experimental data; Ins–instance segmentation; Cls–classification; \# Ins.–number of annotated instances.
}
\centering
\begin{tabular}{lccccccc}
\toprule
Dataset & Source & \# Images & Label & \# Ins. & Scene & Text & Available \\
\midrule
EMPS~\cite{emps} & Literature\textsuperscript{†} & 465 & Ins & 11596 & Diverse & $ \times $ & \checkmark \\
Aversa \etal~\cite{aversa2018first} & Experiment & 22000 & Cls & - & Diverse & $ \times $ & \checkmark \\
Stuckner \etal~\cite{stuckner2022microstructure} & Mixed & 110{,}861 & Cls & - & Diverse & $ \times $ & $ \times $ \\
Okunev \etal~\cite{okunev2020nanoparticle} & Experiment & 26 & Ins & 5852 & Single & $ \times $ & \checkmark \\
Bin \etal~\cite{shi2022automatic} & Experiment & 237 & Ins & 605 & 3 scenes & $ \times $ & $ \times $ \\
Bals \etal~\cite{bals2023deep} & Experiment & 93 & Ins & - & 3 scenes & $ \times $ & $ \times $ \\
MMSci~\cite{li2024mmsci} & Literature & $\ll 100,000^*$ & - & - & Diverse & \checkmark & \checkmark \\
\midrule
EM3M (Ours) & Experiment & 5,091 & Ins & 2,985,660 & Diverse & \checkmark & \checkmark \\
\bottomrule
\end{tabular}
\vspace{0.5em}
\label{tab:em_datasets}
\end{table}

Finally, we benchmark various instance segmentation methods on EM3M, revealing that models originally designed for natural images~\cite{he2017mask,cai2019cascade,chen2019hybrid,bolya2019yolact,Jocher_Ultralytics_YOLO_2023,wang2020solov2,li2023maskdino,cheng2021mask2former,he2023fastinst} suffer accuracy degradation and become computationally and memory-intensive under extreme instance density.
In contrast, micrograph-specific approaches that rely on flow- or embedding-based intermediate representations remain both accurate and resource efficient~\cite{schmidt2018,chen2023cpp,graham2019hover,horst2024cellvit,stringer2021cellpose}; we group these into a “bottom-up” family.
%Building on these insights, we revisit and improve the flow-based framework and train it on a mixture of synthetic and real micrographs, establishing a strong EM-tailored baseline, termed EMNet. We summarize the contribution of this work as follows:
Building on these insights, we revisit and improve the flow-based framework and train it on a mixture of synthetic and real micrographs, establishing a strong baseline called UniAIMS (\textbf{Uni}fied \textbf{AI}-assisted \textbf{M}icroscopy \textbf{S}egmentation). We summarize the contribution of this work as follows:
\begin{itemize}
\item We propose the EM3M, a large-scale and multimodal EM dataset for microstructural segmentation and generation. It comprises 5,091 high-quality EMs, about 3 million instance segmentation labels, and image-level attribute-disentangled textual descriptions. 
\item We release a text-to-image diffusion model trained on the EM3M to serve as a powerful data augmentation tool. Quantitative evaluation metrics and downstream segmentation tasks validate the effectiveness of the generated data.
\item We benchmark various instance segmentation methods on EM3M, and improve the flow-based framework on both synthetic and real micrographs, establishing a strong EM-tailored baseline that serves as a foundation for future microstructural research.
\end{itemize}

\section{Related Work}
\label{sec:relatedwork}
\subsection{Material EM Dataset}
Tab.\ref{tab:em_datasets} summarizes representative EM datasets used in prior work.
The EMPS dataset~\cite{emps} contains 465 EMs with pixel-level segmentation masks, featuring only particle-type microstructures with low instance density and simple scenes. Although Aversa \etal~\cite{aversa2018first} introduced the first large-scale annotated SEM dataset comprising 22,000 images, and Stuckner \etal~\cite{stuckner2022microstructure} used over 100,000 EMs for pre-training, their datasets provide only image-level labels without pixel-level annotations. Additionally, several studies~\cite{okunev2020nanoparticle, bals2023deep, shi2022automatic} still rely on datasets too small for comprehensive instance‑level learning. Other recent efforts draw on EM images extracted from scientific literature, such as MMSci~\cite{li2024mmsci} and OmniSci~\cite{tao2026omniscience}. This dataset contains post-processed images tailored for publication figures, exhibiting reduced realism and lacking raw structural complexity. Moreover, given the scarcity of manually annotated EM datasets, a number of studies~\cite{lin2022deep,lopez2022nanoparticle,mill2021synthetic,vagenknecht2023deep} have explored synthetic data generation to alleviate the annotation burden and improve model performance. However, they typically focus on simplistic, single-material scenarios with limited structural diversity and fail to capture the high-density, heterogeneous microstructural complexity found in real-world EMs.

\subsection{Instance Segmentation} 
Instance segmentation aims to delineate and distinguish individual object instances within an image. Detection-based two-stage methods, such as Mask R-CNN~\cite{he2017mask} and its variants~\cite{cai2019cascade, chen2019hybrid}, are intuitive but often degrade in crowded or highly overlapping scenes. 
In contrast, one-stage methods~\cite{bolya2019yolact, wang2020solov2, lyu2022rtmdet, Jocher_Ultralytics_YOLO_2023} prioritize efficiency by directly predicting instance masks in a single pass, often at the cost of mask quality. More recently, query-based transformers~\cite{cheng2021mask2former,li2023maskdino,he2023fastinst} have achieved state-of-the-art performance by reformulating the task as a direct set prediction problem. However, their reliance on complex bipartite matching for label assignment and a fixed number of learned object queries make them both computationally intensive and difficult to optimize, especially in scenarios with extremely high instance counts.
To address this scalability challenge, a fundamentally different bottom-up, proxy-based paradigm has proven highly effective, particularly for dense imagery in biological microscopy~\cite{schmidt2018, chen2023cpp,  graham2019hover, stringer2021cellpose, jiang2025mcbl}. These methods predict intermediate pixel-level proxies, such as the geometric polygons in StarDist~\cite{schmidt2018} and CPP-Net~\cite{chen2023cpp}, or vector fields like the distance maps in HoverNet~\cite{graham2019hover, horst2024cellvit} and the gradient flows in Cellpose~\cite{stringer2021cellpose, pachitariu2025cellpose}, which are then used to group pixels into individual instances. This bottom-up design bypasses the memory-intensive proposal matching overhead during training, and hence scales better in extreme instance-density scenarios.
%This bottom-up design avoids per-object processing overhead (e.g., NMS and learnable object queries) and hence scales better in extreme instance-density scenarios.
Lastly, interactive methods like SAM~\cite{kirillov2023segment} excel at user-guided segmentation, but their reliance on prompts makes them less effective for fully automatic segmentation in dense scenes.

\section{EM3M Dataset}
\subsection{Data Engine}
\label{data_engine}
\noindent
\textbf{Data collection and processing.}\quad We constructed EM3M by aggregating real-world EMs sourced from three primary channels: (1) high-quality, original EMs from academic collaborations, (2) automated web crawling, and %(3) the annotation of 300 challenging images from Aversa \etal~\cite{aversa2018first}.
(3) 300 representative images selected from Aversa \etal~\cite{aversa2018first} and  newly annotated for instance segmentation. 
The consolidated data underwent a rigorous curation pipeline, including perceptual hash deduplication and a quality filtering process. This process discarded low-resolution images (\textless200px) and employed a GPT-based~\cite{achiam2023gpt} assessment to remove severely blurred images where any embedded text was illegible. After curation, the final dataset comprises approximately 84\% images from academic collaborations, 10\% from web crawling, and 6\% from the public dataset.

Specifically, for training our text-to-image diffusion model, we performed an additional data sanitation step to remove textual artifacts, which could otherwise cause hallucinations and degrade image quality. To eliminate embedded metadata that introduces such artifacts, we trained a YOLO11~\cite{Jocher_Ultralytics_YOLO_2023} model to detect and remove these regions, ensuring the training data contains only pure microstructural content.

\noindent
\textbf{Structured description generation.}\quad To enable fine-grained and structured descriptions of EMs, we designed a descriptive framework based on nine largely disentangled dimensions that characterize the key aspects of a micrograph. These dimensions include metadata (material and microscopy type), visual appearance attributes (color configuration), and microstructural characteristics (morphology, density, distribution, layering in Z-axis, surface texture, and diameter distribution, as shown in Fig.~\ref{fig:sample_image}). Subsequently, we developed a semi-automated annotation pipeline leveraging Gemini~\cite{comanici2025gemini} and GPT. Each model independently extracts nine image attributes into a unified JSON schema, and then cross‑validates the peer’s output to identify inconsistent predictions. Detailed prompts are provided in the supplementary material Sec.~\ref{sec:prompt}. Subsequently, all outputs undergo a mandatory human review. Discrepancies identified between the models are resolved by materials science experts to ensure high-fidelity, attribute-decoupled annotations.

\begin{table}[ht]
    \centering
    \caption{Inter-annotator agreement and evaluation metrics across stages.}
    \label{tab:agreement_and_evaluation}
    \scriptsize
    \setlength{\tabcolsep}{3pt} 
    
    \begin{tabular}{l ccc ccc}
        \toprule
        \multirow{2.5}{*}{Metric} & \multicolumn{3}{c}{\textbf{Inter-annotator agreement}} & \multicolumn{3}{c}{\textbf{Evaluation across stages (vs. GT)}} \\
        
        \cmidrule(lr){2-4} \cmidrule(lr){5-7} 
         & A vs B & B vs C & A vs C & Pseudo Label & Peer Review & Final Validation \\
        \midrule
        KA & 0.823 & 0.810 & 0.816 & 0.695 & 0.801 & 0.818 \\
        PQ & 0.795 & 0.779 & 0.783 & 0.553 & 0.738 & 0.784 \\
        \bottomrule
    \end{tabular}
\end{table}

\noindent
\textbf{Segmentation annotation and quality assessment.}\quad 
We adopted an iterative, human-in-the-loop workflow: domain experts first annotated a high-quality seed set to train an initial segmentation model. To maximize the model's performance and the quality of its predictions in each iteration, we leveraged a diffusion model to augment the training data with diverse, realistic samples. This enhanced model then generated pseudo-labels for the remaining data, which were subsequently refined by trained annotators. To ensure label quality, all annotations underwent a rigorous, multi-stage quality control pipeline, including peer review and final validation by a materials science expert. Additional details are available in the supplementary material Sec.~\ref{sec:annotation}. 
To rigorously validate annotation reliability and the efficacy of this pipeline, we established a stratified subset ($N=300$). This subset was independently re-annotated by three annotators to serve as a benchmark. We measured inter-annotator agreement and assessed the reduction of model bias using pixel-level Krippendorff’s Alpha (KA)~\cite{amgad2022nucls} and Panoptic Quality (PQ). As shown in Tab~\ref{tab:agreement_and_evaluation}, the results indicate consistency among experts, and the steady metric improvements observed across pipeline stages confirm that our protocol effectively mitigates model bias and guarantee expert-level annotation quality.

\subsection{Dataset Analysis}
\label{sec:dataset_analysis}
Through the multi-stage pipeline described above, we constructed the EM3M dataset, comprising 5,091 EMs with 2,985,660 instance-level segmentation annotations. The dataset is randomly divided into 4,128 training images and 963 test images.

\begin{figure}[t]
\centering
\includegraphics[width=0.95\columnwidth]{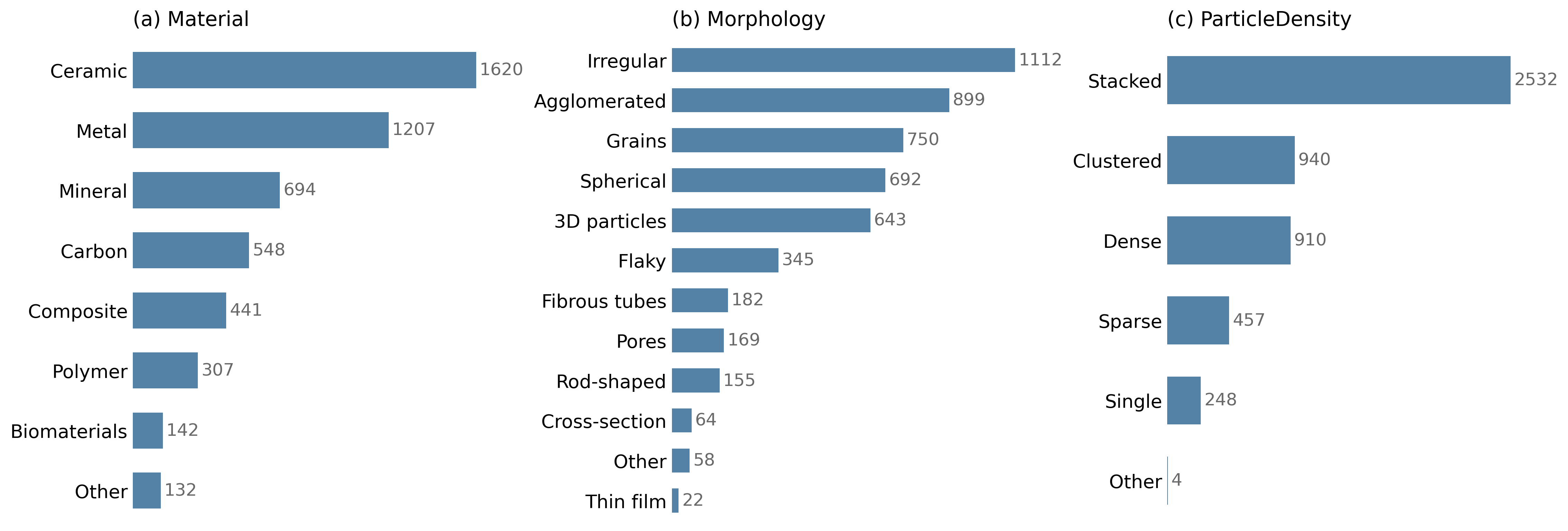}
\caption{Statistical distribution of three representative attributes from EM3M, illustrating its diversity: (a) Material categories, (b) Morphologies, and (c) Particle densities.}
\label{fig:attri_distribution}
\end{figure}

\begin{figure}[t]
\centering
\includegraphics[width=0.85\columnwidth]{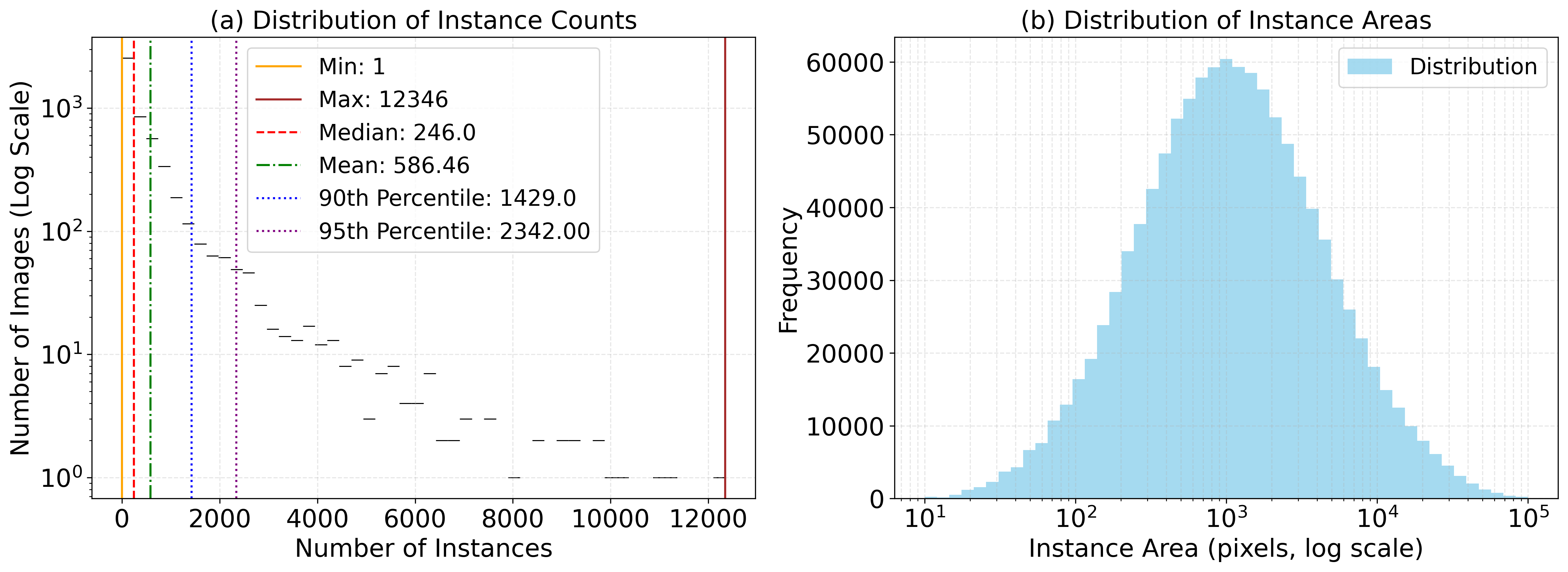}
\caption{Statistical distribution of instances in EM3M (log scale). (a) Distribution of instance counts per image. (b) Distribution of instance areas.}
\label{fig:ins_distribution}
\end{figure}

\noindent
\textbf{Comparison with other EM datasets.}\quad As summarized in Tab.~\ref{tab:em_datasets}, previous EM datasets either focus on classification tasks with a large number of images but no instance-level labels, or provide instance segmentation annotations for a very limited number of images and scenes. While MMSci offers a large-scale collection with textual data, it is sourced from literature, and it lacks precise instance-level annotations. Our EM3M dataset uniquely addresses these limitations by providing the first large-scale collection of nearly 3 million instance segmentation labels derived directly from raw experimental images. Furthermore, it is the only dataset that combines high-density instance labels with structured textual descriptions, providing a crucial resource for developing and evaluating future models for both complex segmentation and multimodal understanding in materials science.

\noindent
\textbf{Universality of EM3M.}\quad We visualized three key attributes—material categories, microstructure morphology, and density—by applying a coarse categorical scheme to accommodate the dataset’s broad coverage and aggregating the corresponding frequencies, as shown in Fig.~\ref{fig:attri_distribution}. Despite a naturally long‐tail distribution across individual categories, the combined coverage encompasses inorganic and organic materials, a wide spectrum of particle shapes (from highly irregular aggregates to well‐defined geometries), and varying degrees of feature density (from isolated particles to heavily overlapped clusters). This breadth of diversity underpins the dataset’s value for developing robust computer‐vision models across challenging EM scenarios. Additional attribute visualizations are available in the supplementary material Sec.~\ref{sec:attri_vis}.

\noindent
\textbf{Instance-level complexity and scale variation.}\quad As shown in Fig.~\ref{fig:ins_distribution}, instance-level statistics reveal highly heterogeneous and challenging scene complexity. The number of annotated microstructures per image spans over four orders of magnitude: while many micrographs contain a few hundred microstructures, a non-negligible portion exhibits extreme crowding (over 1,000 instances), posing significant challenges for segmentation algorithms. Similarly, the distribution of instance areas—from ten-pixel segments to over $10^5$ pixels on a logarithmic scale—indicates substantial scale variation. Additional statistical visualizations are provided in the supplementary material Sec.~\ref{sec:distribution}.

\section{Methodology}
\subsection{Benchmarking Baseline}
To identify a suitable framework for microstructural analysis, we benchmark representative instance segmentation paradigms on EM3M. Our empirical results (Sec.~\ref{sec:Instexp}) demonstrate that flow-based approaches, exemplified by Cellpose~\cite{stringer2021cellpose, pachitariu2025cellpose}, provide a more stable and competitive reference than detection- and query-based methods under extreme instance density. This advantage stems from their pixel-wise flow representation, which directly aggregates local gradients into object instances and is less constrained by proposal or query mechanisms in crowded scenarios. A summary of the principle is in the supplementary material Sec.~\ref{sec:cellpose_principle}.

Guided by this empirical observation, we construct a strong baseline, termed UniAIMS, based on the Cellpose flow formulation. To better match the characteristics of EM3M, we introduce several practical adaptations: we remove the built-in size model to accommodate substantial scale variation, adopt scale-robust data augmentation, enable distributed training with large-resolution inputs, and modernize the backbone for stronger feature representation. These adjustments are implementation-level refinements tailored to large-scale EM data, with further details provided in the supplementary material Sec.~\ref{sec:emnet}.

\subsection{SDXL Finetune}
Diffusion models~\cite{ddpm, sd, zhu2025fmri2ges,  zhao2026resilphaseplugandplayphasemapping, huang2026exposurebiasalleviatedirectional}, especially Stable Diffusion and its advanced variant SDXL, have recently achieved remarkable progress in text-guided image generation. Methods such as Textual Inversion~\cite{null} and DreamBooth~\cite{dreambooth} further adapt these pretrained models to specific subjects. Moreover, LoRA-based methods~\cite{lora} significantly reduce the computational cost of fine-tuning, enabling efficient content and style customization. 

In this work, we adopt the LoRA training scheme and introduce several learnable tokens to represent different categories within EM3M. Each token is initialized with a rare word and optimized during LoRA fine-tuning, allowing it to better capture category-specific representations. This training process is guided by a carefully prepared dataset of paired images and structured descriptions. On the image side, micrographs from EM3M are processed to remove any regions containing text, ensuring a clean, high-quality dataset free from textual artifacts (see Sec.~\ref{data_engine} for details). On the text side, each image is paired with a structured prompt constructed from a set of quasi-orthogonal attributes (see Fig.~\ref{fig:sample_image}). The prompt follows the consistent template shown below:

\begin{tcolorbox}[colback=gray!5!white, colframe=gray!40!white, boxrule=0.4pt, arc=1mm, left=3pt, right=3pt, top=3pt, bottom=3pt]
\ttfamily\scriptsize
\textless microscopy\_type\textgreater{} of \textless material\textgreater: \textless morphology\textgreater.
\textless texture\textgreater. \textless density\textgreater. \textless distribution\textgreater. 
\textless layering\textgreater. \textless diameters\textgreater. \textless color\textgreater.
\end{tcolorbox}
%Specifically, the fine-tuning dataset consists of paired images and structured descriptions. Real images from EM3M are processed to remove any text regions, ensuring a clean, high-quality generative dataset (see Sec.\ref{data_engine} for details). This step prevents the model from learning unwanted biases tied to embedded text, which could otherwise lead to anomalous outputs.  In parallel, each image is paired with a structured prompt, constructed from a set of quasi-orthogonal attributes, as shown in Fig.\ref{fig:sample_image}. This prompt follows a consistent template, as shown below:

%By combining these descriptions with LoRA-based fine-tuning of both adapter weights and word embeddings, our SDXL gains strong attribute-level control. The fine-tuned model can generate faithful variations of existing samples and is also capable of synthesizing novel microstructures by recombining these descriptive attributes.

The structured format is crucial for achieving fine-grained control over the synthesis process. By leveraging LoRA-based fine-tuning of both adapter weights and word embeddings, our model can generate plausible variations when sampling attributes based on their observed correlations, resulting in microstructures that adhere to realistic material properties. Furthermore, this attribute-level control allows for the exploratory synthesis of novel images by randomly recombining attributes, which can generate diverse, sometimes physically implausible, samples potentially useful for enriching the data distribution. Visualization examples of controllable synthesis, as well as the impact of synthetic data on instance segmentation performance, are provided in Sec.~\ref{sec:exp}.

\begin{table}[t]
\caption{Baseline performance comparison of various instance segmentation methods on the sparse subset and the full dataset.}
\centering
\scriptsize
\begin{tabular}{llcccccc}
\toprule
\multirow{2}{*}{Class} &\multirow{2}{*}{Method} & \multirow{2}{*}{\centering Backbone} & \multirow{2}{*}{\centering Params} &\multicolumn{2}{c}{Sparse} & \multicolumn{2}{c}{All} \\
%\cmidrule(lr){3-10}
        &    &  &  & $mAP@0.5$ & $PQ@0.5$ & $mAP@0.5$ & $PQ@0.5$  \\
\midrule
%\multirow{3}{*}{Field-based Models} \\
\multirow{3}{*}{Two-stage}     & Mask R-CNN~\cite{he2017mask}           &  ResNeXt101 & 101M & 0.620 & 0.610 & - & - \\
                               & Cascade R-CNN~\cite{cai2019cascade}    &  ResNeXt101 & 135M & 0.624 & 0.622 & - & - \\
                               & HTC~\cite{chen2019hybrid}              &  ResNeXt101 & 137M & 0.497 & 0.538 & - & - \\
\midrule
\multirow{3}{*}{One-stage}   
                               & YOLACT~\cite{bolya2019yolact}         &  ResNeXt101 & 93M & 0.538 & 0.537 & - & - \\
                               & YOLO11-seg~\cite{Jocher_Ultralytics_YOLO_2023} & ResNeXt101 & 116M & 0.619 & 0.605 & - & - \\
                               & SOLOv2~\cite{wang2020solov2} & ResNeXt101 & 113M & 0.339 & 0.454 & - & - \\
\midrule
\multirow{3}{*}{Query-based}
                               & FastInst~\cite{he2023fastinst} & ResNeXt101 &   93M   &  0.332     &   0.466    & - & - \\
                               & Mask2Former~\cite{cheng2021mask2former} & Swin-Base & 107M & 0.296 & 0.301 & - & - \\
                               & MaskDINO~\cite{li2023maskdino} & Swin-Base & 115M & 0.395 & 0.515 & - & - \\
                               
\midrule
\multirow{5}{*}{Bottom-up}   
                               & Stardist~\cite{schmidt2018}           &    ViT-Base    &  93M    & 0.670 & 0.559 & 0.668 & 0.582 \\
                               & CPP-Net~\cite{chen2023cpp}        &     ViT-Base    &   96M   & 0.663 & 0.602 & 0.647 & 0.528 \\
                              & HoverNet~\cite{graham2019hover} & ViT-Base & 104M & 0.620 & 0.549 & 0.594 & 0.554 \\
                              & CellViT~\cite{horst2024cellvit}    &    ViT-Base       &   131M   & 0.608 & 0.625 & 0.672 & 0.600 \\
                              & Cellpose~\cite{pachitariu2025cellpose}    & ViT-Base & 93M & 0.650 & 0.692 & 0.711 & 0.691 \\
                              %& Omnipose~\cite{cutler2022omnipose} & ViT-Base & 96M & & & 0.646 & 0.646 \\
\midrule
                              & UniAIMS (Ours)         & ViT-Base  &  93M & \textbf{0.818} & \textbf{0.769} & \textbf{0.800} & \textbf{0.704} \\
\bottomrule
\end{tabular}%
\label{exp:baseline}
\end{table}

\begin{figure}[t]
    \centering
    \includegraphics[width=0.98\columnwidth]{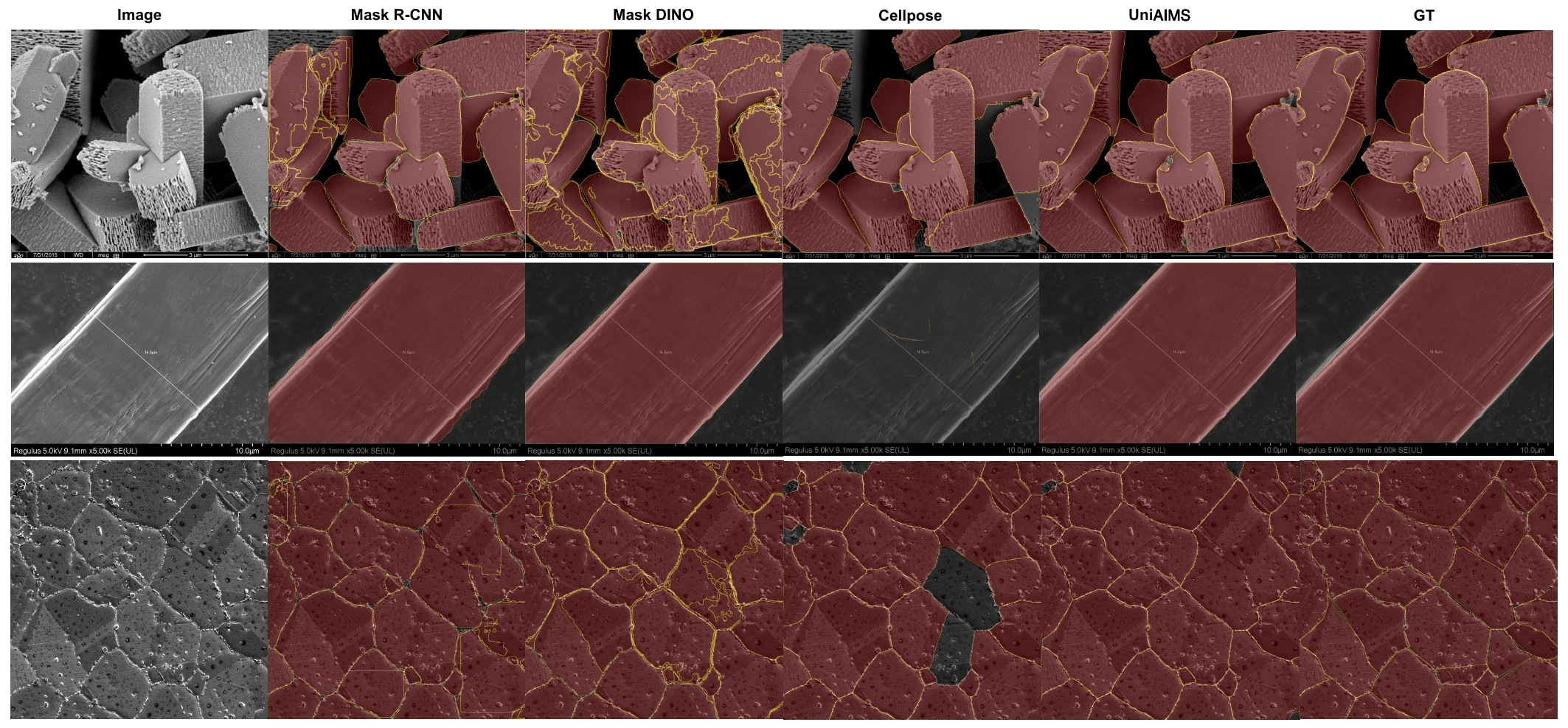} 
    \caption{Ground truth and predictions of different methods on the sparse subset.}
    \label{fig:display}
\end{figure}

\begin{figure}[t]
\centering
\includegraphics[width=0.95\columnwidth]{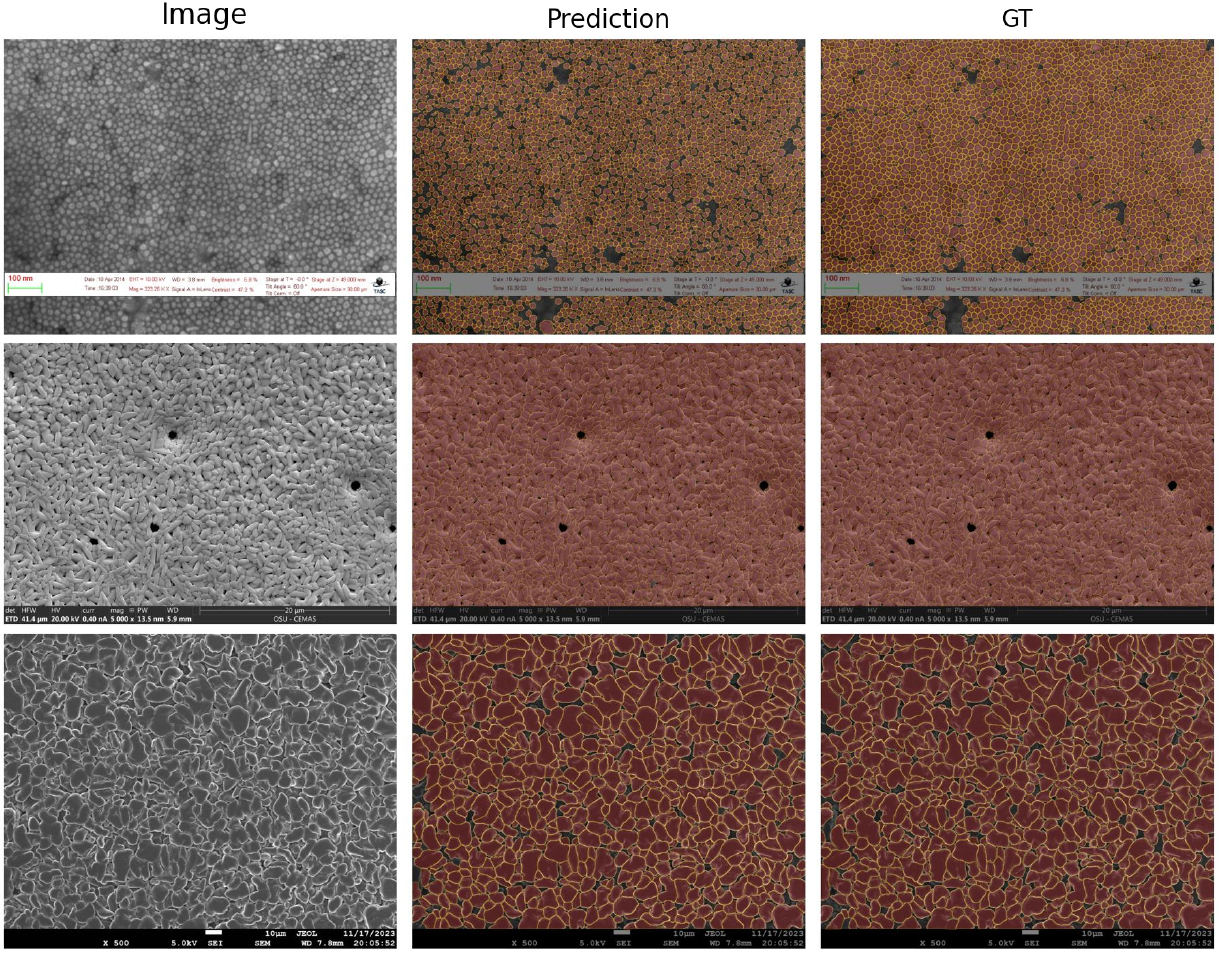} 
\caption{Visualization of our prediction result for images with dense instances.}
\label{fig:dense}
\end{figure}

%%%%%%%%%%%%%---------------
\begin{figure}[tb]
  \centering
  \begin{subfigure}{0.45\linewidth}
    \includegraphics[width=\linewidth]{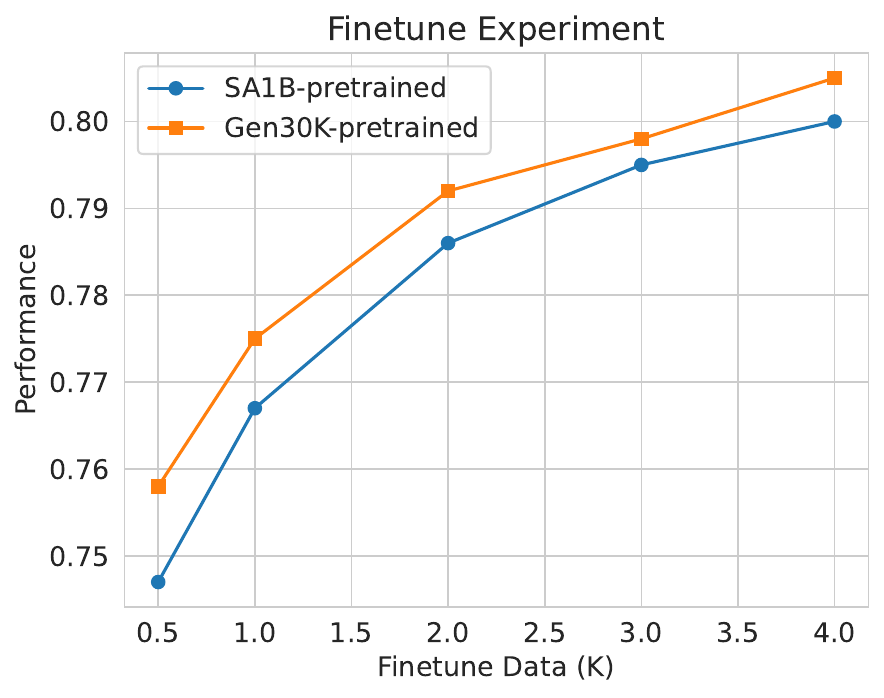}
    \caption{Comparison of fine-tuning performance.}
    \label{fig:finetune}
  \end{subfigure}
  \hfill
  \begin{subfigure}{0.45\linewidth}
    \includegraphics[width=\linewidth]{{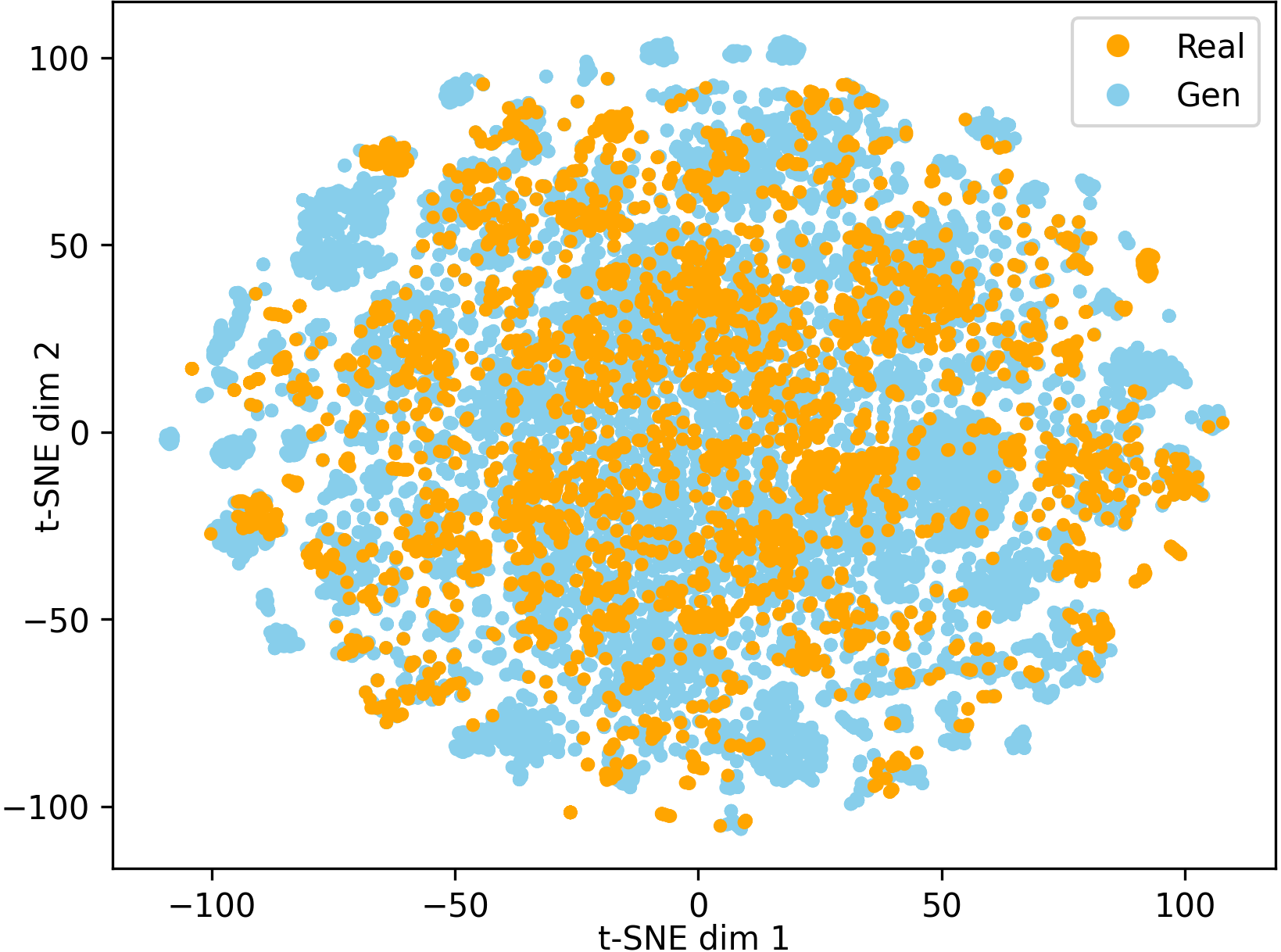}}
    \caption{t-SNE visualization of data distribution.}
    \label{fig:tsne}
  \end{subfigure}
  \caption{Evaluation of the generated dataset quality. (a) Comparison of models pretrained on SA1B versus our 30K generated images (Gen30K). (b) t-SNE visualization showing the substantial feature space overlap between real (orange) and generated (blue) data, indicating high domain fidelity.}
  \label{fig:eval_gen}
\end{figure}

\section{Experiment}
\label{sec:exp}
\subsection{Instance Segmentation}
\label{sec:Instexp}
\noindent
\textbf{Comparison methods.}\quad For our benchmark, we select representative instance segmentation methods, including four major architectural paradigms. We evaluate classic detection-based approaches, including two-stage models that follow a ``detect-then-segment'' strategy (Mask R-CNN~\cite{he2017mask}, Cascade R-CNN~\cite{cai2019cascade}, HTC~\cite{chen2019hybrid}) and faster one-stage models (YOLACT~\cite{bolya2019yolact}, YOLO11-seg~\cite{Jocher_Ultralytics_YOLO_2023}, SOLO\-v2~\cite{wang2020solov2}). We also include query-based architectures (Mask2Former~\cite{cheng2021mask2former}, Mask\-DINO~\cite{li2023maskdino}, FastInst~\cite{he2023fastinst}), which treat segmentation as a direct set prediction task. Finally, to specifically address the challenges of dense imagery, we assess a group of bottom-up, proxy-based methods, including StarDist~\cite{schmidt2018} (polygon-based) and its variant CPPNet~\cite{chen2023cpp}, HoverNet~\cite{graham2019hover} and CellViT~\cite{horst2024cellvit} (embedding-based), and Cellpose (flow-based)~\cite{stringer2021cellpose}, which predict intermediate pixel-level representations before reconstructing instances.

\noindent
\textbf{Implementation details.}\quad To ensure rigor and reproducibility, all benchmarked methods are implemented using their official public codebases, primarily through the MMDetection~\cite{chen2019mmdetection} and Detectron2~\cite{wu2019detectron2} framework, with specialized repositories integrated as needed.
%The training is conducted on 4 NVIDIA 4090D GPUs with a batch size of 8. 
Following CellViT~\cite{horst2024cellvit}, we equip all bottom-up methods with a SAM-pretrained ViT-B backbone to ensure fair comparison. Correspondingly, other methods are benchmarked using backbones of comparable scale, such as ResNeXt\-101~\cite{resnext} and Swin-B~\cite{liu2021swin}. 
Given that detection-based and query-based architectures struggle with the extreme instance densities found in our dataset, we benchmarked them exclusively on a sparse subset containing images with fewer than 100 instances. Conversely, all bottom-up methods, including UniAIMS, were evaluated on the full dataset to demonstrate their scalability.
A detailed justification for this evaluation strategy, along with other implementation specifics, is provided in the supplementary material Sec.~\ref{sec:implementation}.

\noindent
\textbf{Evaluation metrics.}\quad We evaluate instance segmentation performance using mean Average Precision (mAP) for mask detection accuracy and Panoptic Quality (PQ)~\cite{kirillov2019panoptic} at 0.5 IoU for combined segmentation and recognition quality. Following prior work in bioimage analysis~\cite{schmidt2018,stringer2021cellpose,graham2019hover}, we report AP as $\text{AP} = \frac{|TP|}{|TP| + |FP| + |FN|}$, where predictions with IoU above a threshold $T$ are counted as true positives. Mean AP is obtained by averaging AP across images. This definition differs from the COCO-style AP, but has become a standard practice in microscopy instance segmentation. 

\noindent
\textbf{Main results.}\quad Tab.~\ref{exp:baseline} presents a comprehensive quantitative comparison of representative instance segmentation paradigms on EM3M, and the qualitative examples in Fig.~\ref{fig:display} corroborate these trends while revealing characteristic failure modes. Additional visualizations and analysis are provided in the supplementary material Sec.~\ref{sec:prediction}.

Query-based models generally underperform on our benchmark: for instance, Mask DINO attains a mAP of only 0.395 on the sparse split, and frequently produces fragmented or non-cohesive masks that fail to separate adjacent instances in cluttered regions (Fig.~\ref{fig:display}, column 3). For microstructures with complex surface topographies and high-frequency texture, the direct set-prediction strategy in these architectures may be less effective at recovering continuous, fine-grained boundaries that are important for morphological analysis.

Detection-based methods (e.g., Mask R-CNN, Cascade R-CNN) remain competitive, which indicates their utility for simpler scenes with a lower density of objects. However, bottom-up approaches show a consistent advantage overall. Notably, Cellpose achieves a strong balance between instance recall and mask fidelity, as evidenced by the relatively small gap between its mAP and PQ scores. This motivated our decision to adopt its flow-based mechanism as the foundation for our model. A key vulnerability of bottom-up pipelines, however, is their iterative pixel-grouping post-processing, which can accumulate errors and occasionally lead to the complete omission of very large instances (Fig.~\ref{fig:display}, row 2).

Building upon the flow-based paradigm, our proposed UniAIMS mitigates these limitations by improving multi-scale processing and feature representation. The resulting model achieves a state-of-the-art mAP of 0.818 on the sparse subset and maintains robust performance (0.800 mAP) on the full, more challenging dataset. The quantitative and qualitative evidence strongly indicates that proxy-based designs, particularly when augmented for scale robustness, are exceptionally well-suited for dense, texture-rich electron micrographs. The superiority of our method is most pronounced in highly dense scenarios. As visualized in Fig.~\ref{fig:dense}, UniAIMS delivers highly accurate segmentation for densely packed and irregularly shaped instances, with predictions that closely align with the ground truth, confirming the effectiveness of our approach in addressing these unique challenges.

%As shown in Tab.~\ref{exp:baseline}, proxy-based methods consistently outperform traditional anchor-based or anchor-free methods on our datasets. Anchor-based methods, such as Mask R-CNN and HTC, struggle more or even fail in dense scenarios due to computational limitations in NMS and bounding box proposals. We partition our dataset into sparse (fewer than 100 instances) and dense (more than 100 instances) data. All anchor-based methods are solely trained and evaluated on the sparse subset. These experiments demonstrate the superior capability of flow-based methods, particularly UniEM-Net, in handling high-density microstructures, likely attributable to their effective use of 2D fields for instance separation in crowded scenes. Visually, this performance gap is more pronounced, as shown in Fig.~\ref{fig:display}, which displays the segmentation results of some methods on the sparse dataset.

%Our method's superiority is most evident in dense scenarios. Fig.~\ref{fig:dense} presents a visual comparison of our prediction results with the ground truth for images with dense instances. It is clear that UniEM-Net provides highly accurate segmentation for densely packed and irregularly shaped instances, closely matching the ground truth. This visual evidence, combined with the quantitative results in Tab.~\ref{exp:baseline}, confirms the effectiveness of our approach in addressing the unique challenges posed by electron micrographs. More visualization results are provided in the supplementary material Sec.~\ref{sec:prediction}.

\begin{table}[t]
\centering

% ================= LEFT: Table 4 =================
\begin{minipage}[t]{0.48\textwidth}
\centering
\caption{Performance metrics for models trained on varying amounts of real data, generated data, and their combination.}
\label{exp:mount}
\small
\begin{tabular}{lcccc}
\toprule
Modality & Amount & $mAP@0.5$ & $PQ@0.5$ \\
\midrule
\multirow{4}{4em}{Real} & 1 K & 0.767 & 0.682  \\
                        & 2 K & 0.786 & 0.696  \\
                        & 3 K & 0.795 & 0.700  \\
                        & 4 K & 0.800 & 0.704 \\
\midrule
\multirow{4}{4em}{Gen}  & 5 K  & 0.760 & 0.669  \\
                       & 10 K & 0.771 & 0.676  \\
                       & 20 K & 0.785 & 0.686  \\
                       & 30 K & 0.791 & 0.690  \\
\midrule
Gen+Real & 30+4 K & \textbf{0.809} & \textbf{0.707}  \\
\bottomrule
\end{tabular}
\end{minipage}
\hfill
% ================= RIGHT: Table 5 & 6 =================
\begin{minipage}[t]{0.48\textwidth}
\centering

% ---------- Table 5 ----------
\begin{minipage}[t]{\textwidth}
\centering
\caption{UniAIMS Performance on external datasets.}
\label{tab:other_dataset}
\small
\begin{tabular}{lccc}
\toprule
Dataset & Reported & Zero-shot & Finetune \\
\midrule
EMPS\cite{emps} & 0.92  & 0.89 & 0.92 \\
Okunev\cite{okunev2020nanoparticle} & 0.75 & 0.81 & 0.84 \\
\bottomrule
\end{tabular}
\end{minipage}

\vspace{1.0em}

% ---------- Table 6 ----------
\begin{minipage}[t]{\textwidth}
\centering
\caption{Generative metrics between the real datasets and the generative dataset.}
\label{tab:gen}
\small
\setlength{\tabcolsep}{5pt}
\begin{tabular}{lccc}
\toprule
Metric & Real & Base & Ours \\
\midrule
FID$(\downarrow)$ & 28.06 & 213.16 & 34.03 \\
Style$(\uparrow)$ & 0.985 & 0.669 & 0.958 \\
\bottomrule
\end{tabular}
\end{minipage}

\end{minipage}

\end{table}

\noindent
\textbf{Varying amounts of data.}\quad 
Tab.~\ref{exp:mount} demonstrates a clear positive correlation between the amount of real training data and model performance, with mAP steadily increasing from 0.767 (1K images) to 0.800 (4K images).
A similar trend is observed for generated data, where performance also increases with data volume but shows diminishing gains at larger scales.
Training solely on 30K generated images (0.791 mAP) yields performance approaching that of 3K real images (0.795 mAP), underscoring the high quality of our synthesized data.
The best results, 0.809 mAP and 0.707 PQ, are obtained when real and generated data are jointly used, confirming that our synthetic data serves as a powerful and effective tool for data augmentation.

\noindent
\textbf{Pretraining with generated data.}\quad 
To further assess the utility of the synthesized dataset, we pretrain our model on 30K generated images and finetune it with varying amounts of real data, as illustrated in Fig.~\ref{fig:finetune}.
Compared with the SA1B-pretrained~\cite{kirillov2023segment} counterpart, the model pretrained on our generated data consistently achieves superior performance across all real-data scales, demonstrating the strong transferability and pretraining potential of our synthetic dataset.

\noindent
\textbf{Cross-Dataset Evaluation and Benchmark Saturation.}\quad We evaluated UniAIMS on external public datasets (EMPS~\cite{emps}, Okunev \etal~\cite{okunev2020nanoparticle}). As shown in Tab.~\ref{tab:other_dataset}, UniAIMS achieves strong zero-shot performance, significantly outperforming existing baselines even without domain-specific training. Moreover, with fine-tuning, performance on these datasets saturates at near-perfect levels. This performance ceiling suggests that prior datasets may be insufficiently challenging for state-of-the-art models due to limited structural complexity. In contrast, EM3M poses substantially greater difficulty, underscoring its necessity as a robust benchmark to drive future advances in EM segmentation.

\begin{figure}[tb]
\centering
\includegraphics[width=0.95\columnwidth]{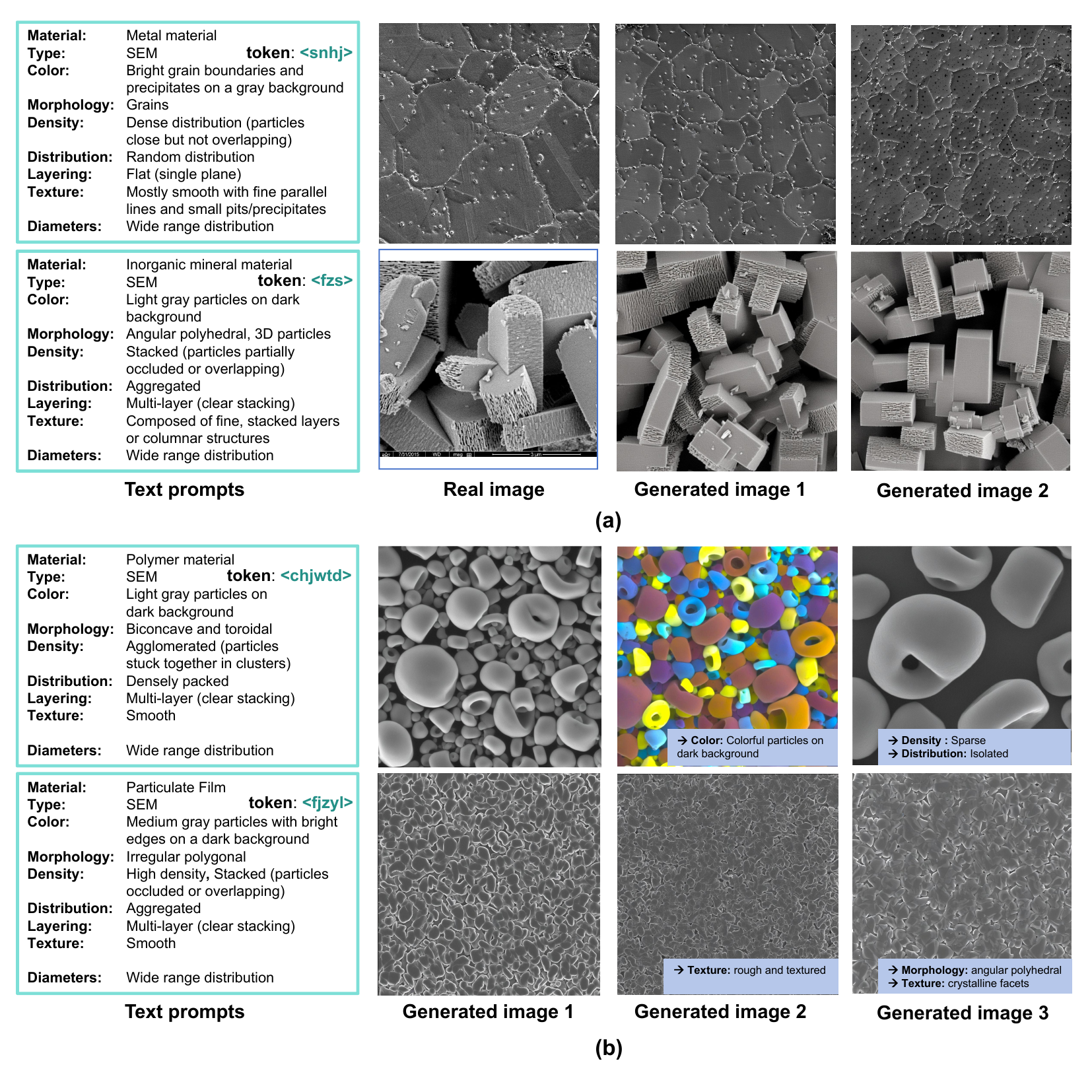} % Reduce the figure size so that it is slightly narrower than the column. Don't use precise values for figure width.This setup will avoid overfull boxes.
\caption{Visual samples of generated data. (a) demonstrates distributional consistency between real images and text-conditioned generated samples, while (b) illustrates novel out-of-distribution examples produced through attribute recombination.}
\label{fig:gen}
\end{figure}

\subsection{Synthesized Data}
\label{SynthesizedData}
This section presents qualitative and quantitative analysis of the generative model (finetuned SDXL), focusing on aspects such as fidelity and distribution consistency.

\noindent
\textbf{Quantitative analysis.}\quad
We quantitatively assess our generative model in Tab.~\ref{tab:gen}, where \textit{Real Data} denotes the processed real micrographs without text regions, and \textit{SDXL (base)} refers to the synthetic dataset produced by the vanilla SDXL model without fine-tuning.
To measure distributional consistency, we adopt the Fréchet Inception Distance (FID)~\cite{fid}. Since the standard FID relies on an ImageNet-trained feature extractor that is suboptimal for EMs, we first establish a domain-specific reference by computing the FID between real images and their spatially augmented counterparts (rotation, scaling, and flipping). This reference value of 28.06 reflects the inherent sensitivity of FID within our domain and serves as a baseline for meaningful comparison. Our fine-tuned model achieves an FID of 34.03, substantially improving over the SDXL baseline (213.16) and approaching the domain reference, indicating strong generative fidelity.
%Tab.\ref{tab:gen} presents the evaluation of consistency between ground truth and synthesized samples, where Real Data represents the processed real data without any text regions, and SDXL(base) represents the synthesized dataset generated by SDXL without fine-tuning. First, we employed FID (Fréchet Inception Distance)~\cite{fid} to measure the overall distributional consistency against the generative training dataset. Real Data achieved the best FID score 28.06. While our FID score of 34.03 is slightly worse, this remains reasonably acceptable given the variations between synthesized samples and training images. 
We further evaluate style similarity by extracting the [CLS] tokens of both real and generated images using a pre-trained DINO-ViT~\cite{dino} and computing their cosine similarity. As shown in Tab.~\ref{tab:gen}, the high Style score (0.958) confirms that our synthesized images closely align with the stylistic distribution of real micrographs. This, combined with the analysis from segmentation experiments, further validates the effectiveness of the synthesized data and its generalization capability for downstream tasks.
%\begin{figure}[t]
%    \centering
%    \includegraphics[width=0.65\columnwidth]{figures/tsne.png}
%    \caption{t-SNE analysis of real and generated datasets.}
%    \label{fig:tsne}
%\end{figure}

\noindent
\textbf{Qualitative analysis.}\quad
Fig.~\ref{fig:tsne} illustrates the t-SNE visualization results comparing the synthesized and real datasets, where the highly overlapping distributions confirm their consistency. Furthermore, visual samples in Fig.~\ref{fig:gen}(a) demonstrate that our model can produce high-quality, realistic EM data. Notably, generative EM samples from different categories show obvious distinctions and exhibit visual attributes highly similar to real samples. Meanwhile, Fig.~\ref{fig:gen}(b) shows the capability of generative model to generate novel, out-of-distribution images via attribute recombination. This further demonstrates the model's ability to learn representations across various categories and attributes. Overall, both qualitative and quantitative analyses confirm the potential of our synthetic dataset to replace and augment real datasets.

\section{Open Problems}
Despite the strong performance of UniAIMS and the scale of EM3M, several important challenges remain unsolved. First, extremely dense micrographs containing heavily clustered and stacked instances continue to be difficult even for flow-based methods. Moreover, while AP/PQ@50 provides a useful benchmark, these metrics may be overly lenient in crowded scenarios and do not fully reflect fine-grained boundary quality. For example, although UniAIMS achieves 0.704 PQ@50, its performance drops substantially to 0.340 under the stricter PQ@90 criterion, indicating considerable room for improvement in precise instance delineation. Second, microstructures with fibrous textures, weak contrast, and ambiguous boundaries remain challenging due to the lack of reliable structural cues for instance separation (Fig.~\ref{fig:main_results}). Third, substantial intra-image scale variation, where particle sizes may span several orders of magnitude within a single micrograph, continues to challenge current architectures despite scale-aware training strategies. Furthermore, a noticeable gap remains between human annotation consistency and model performance: under the same PQ metric, average human agreement reaches 0.786, whereas the best model attains 0.704. Together, these findings suggest that future research should explore more effective dense-scene representations, boundary-aware learning objectives, scale-adaptive architectures, and evaluation protocols that better capture segmentation quality in highly crowded micrographs.

\section{Conclusion}
In this study, we introduce EM3M, a large-scale and multimodal electron microscopy dataset for microstructural segmentation and generation in materials science. Building on this foundation, we release a text-to-image diffusion model that serves as an effective data augmentation engine, with quantitative and downstream results validating the utility of the generated samples. Through comprehensive benchmarking, we demonstrate that an improved flow-based framework performs robustly on synthetic and real micrographs, establishing a strong EM-tailored baseline. Together, EM3M, the UniAIMS baseline, and the generative diffusion model form a unified toolkit that we expect will accelerate progress in automated materials analysis and material-oriented vision–language research.
%We also propose UniAIMS, a strong baseline model that outperforms state-of-the-art methods in microstructural segmentation. Additionally, we release a text-to-image diffusion model trained on this dataset, which serves as a powerful data augmentation tool.
%We believe that the EM3M dataset, the UniAIMS baseline model, and the generative diffusion model will collectively form a comprehensive toolkit, significantly accelerating research in automated materials analysis and material-oriented vision-language models.

\bibliographystyle{splncs04}
\bibliography{main}

\newpage
\setcounter{page}{1}
% \maketitlesupplementary

\section{The Principle of Cellpose}
\label{sec:cellpose_principle}
\subsection{Segmentation via Gradient Flows}
Unlike traditional top-down methods that first detect objects and then segment them, Cellpose~\cite{stringer2021cellpose} reformulates the task by learning a global, holistic representation of each instance's geometry. The core principle is to predict a gradient vector field for all pixels within an instance, where each vector directs towards the instance's geometric center. This gradient field is derived during training by transforming each ground-truth binary mask into a continuous spatial representation. Specifically, a diffusion process is simulated from the boundary of a mask inwards, creating a scalar potential field whose value is highest at the center.

The network is then trained to predict a multi-channel output based on this information. The primary outputs are the x and y components of the normalized gradient field, which form the 2D vector flow. In addition, to distinguish objects from the background, the network also learns to predict a probability map indicating whether each pixel belongs to an instance (foreground). Together, these three channels serve as the complete set of supervision signals for the model.

\subsection{Scale Normalization via the Size Model}
A key architectural choice in the original Cellpose framework is its strategy for handling object scale. The main segmentation network is not designed to be inherently scale-invariant; rather, it is trained on objects that have been rescaled to a consistent, canonical diameter (e.g., 30 pixels). To accommodate this during inference, Cellpose first employs a separate, pre-trained size model to estimate the average object diameter in a given input image. The image is then resized by a factor that aligns this estimated diameter with the canonical diameter before being processed by the main segmentation network. Finally, the resulting instance masks are scaled back to the original image dimensions.

This approach is highly effective when objects within an image share a relatively uniform scale, as is common in many biological microscopy applications. However, its core assumption of limited intra-image scale variance makes it unsuitable for datasets like EM3M, where particle sizes can span orders of magnitude within a single micrograph. This critical limitation directly motivated our decision to remove this component in UniAIMS and instead employ data augmentation (LSJ) to achieve scale robustness.

\subsection{Mask Reconstruction from Predicted Flows}
During the inference stage, the trained model predicts the three output channels defined previously—the 2D vector flow (x and y components) and the foreground probability map—for a given micrograph. Instance masks are then reconstructed through a deterministic process that follows these predicted flows. Starting from foreground pixels identified by the probability map, each pixel is iteratively moved according to the predicted vector field, following a numerical integration scheme akin to a modified Euler method. All pixels that converge to the same spatial sink point where the flow magnitude approaches zero are grouped together to form a single instance. This bottom-up, flow-based approach excels at delineating boundaries between tightly packed or touching objects based on their global shape, making it an exceptionally suitable baseline for the complex microstructures prevalent in the EM3M dataset.

\begin{figure}[tb]
  \centering
  \begin{subfigure}{0.68\linewidth}
    \includegraphics[width=\linewidth]{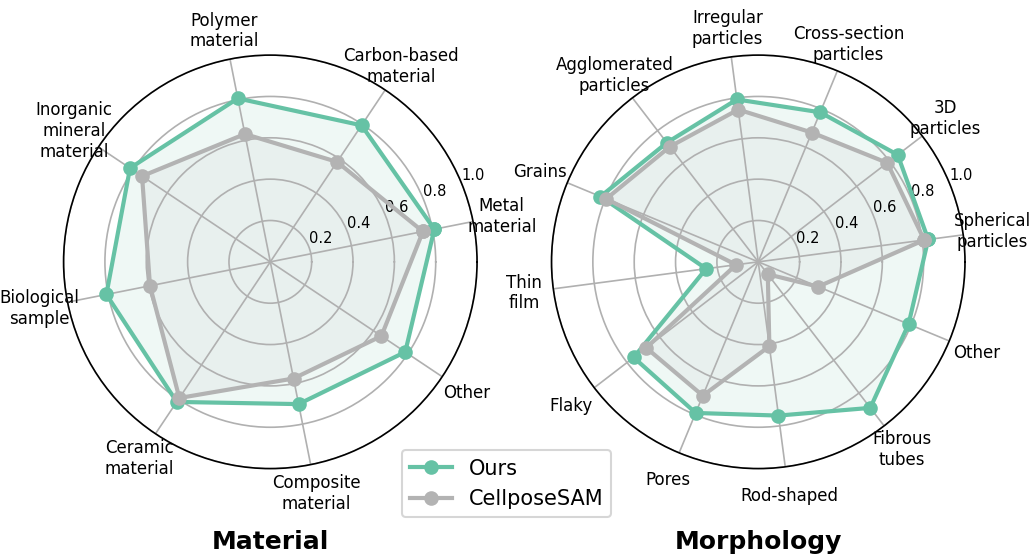}
    \caption{Comparison across diverse data subsets.}
    \label{fig:subset_test}
  \end{subfigure}
  \hfill
  \begin{subfigure}{0.3\linewidth}
    \includegraphics[width=\linewidth]{{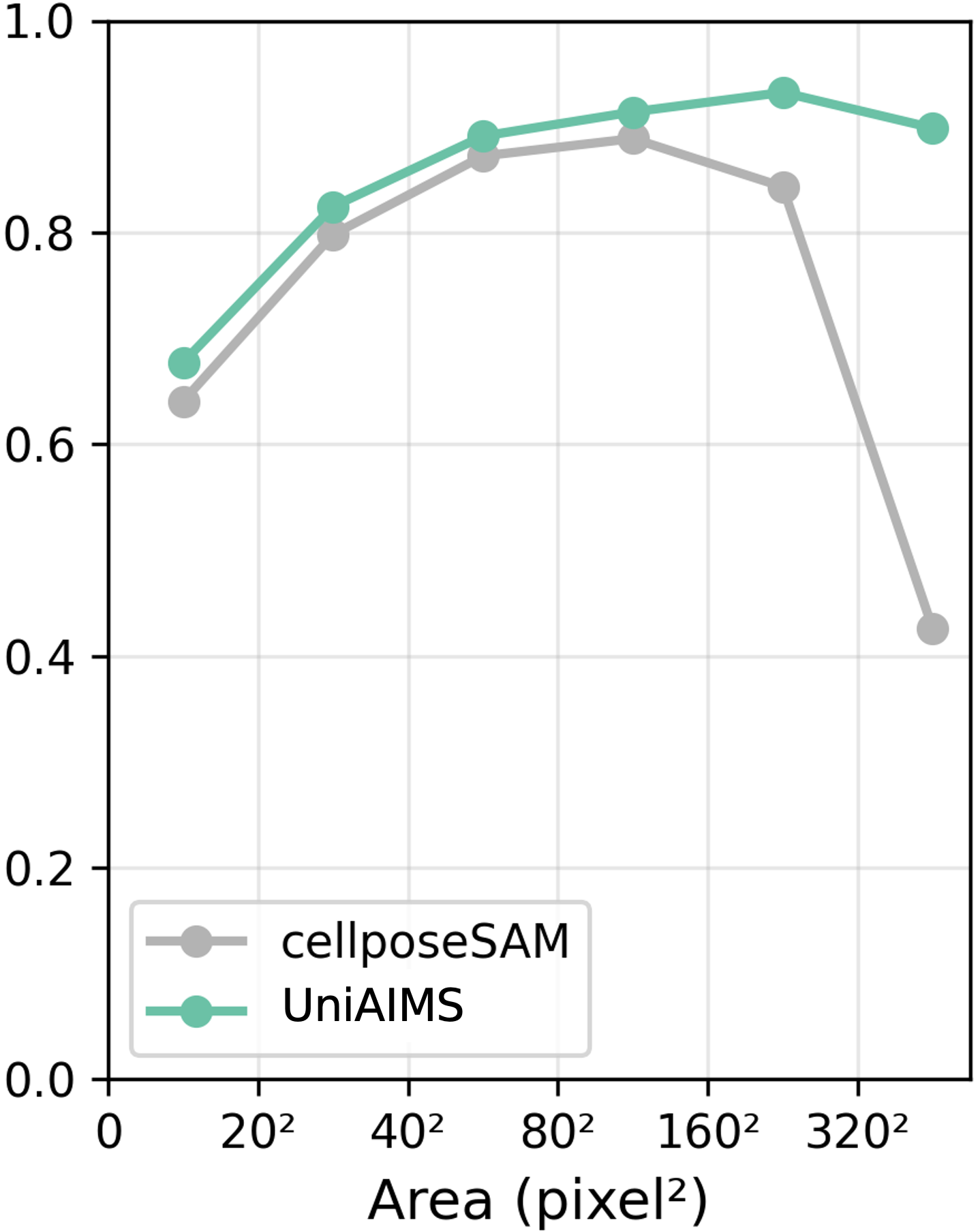}}
    \caption{AP50 on different targe size of EM3M.}
    \label{fig：diff_size}
  \end{subfigure}
  \caption{Performance comparison between UniAIMS and Cellpose}
  \label{fig:comparison}
\end{figure}

\section{Adaption of Cellpose: UniAIMS}
\label{sec:emnet}
To establish a robust performance benchmark on EM3M, we configure a strong baseline termed UniAIMS. Rather than introducing a new segmentation architecture, UniAIMS builds upon the well-established Cellpose~\cite{stringer2021cellpose} flow-based paradigm, which is inherently well suited for delineating dense and crowded microstructures in micrographs.

Specifically, we modified the original implementation in three key aspects. First, we re-engineered the codebase to support multi-GPU distributed training and larger batch sizes and input resolutions, addressing the original implementation’s single-GPU constraint and consequent training inefficiency. Second, recognizing that Cellpose’s  built-in size-model assumes limited within-image scale variation and fails on images with large particle-size diversity, we removed this component and instead employed Large Scale Jittering (LSJ) augmentation to enforce scale robustness. Third, to make LSJ and flow-field synthesis feasible at scale, we implemented a CUDA-accelerated flow computation module that reduces preprocessing overhead and supports real-time augmentation. We also upgraded the backbone to a ViT-Base to strengthen feature representation for diverse microstructures. We stratified EM3M into diverse subsets, and as illustrated in Fig.~\ref{fig:subset_test}, UniAIMS consistently outperforms Cellpose~\cite{pachitariu2025cellpose} across diverse material types and morphological categories, demonstrating stronger robustness to structural diversity and complex particle geometries. Moreover, Fig.~\ref{fig：diff_size} shows that UniAIMS maintains stable performance across varying target sizes, whereas Cellpose degrades significantly on large objects. These results validate that our implementation-level adaptations substantially improve scalability and scale robustness for EM segmentation.

\section{Prompt Design for Structured Description}
\label{sec:prompt}
To facilitate structured description generation, we employed Gemini and GPT to independently extract the nine key attributes from batches of EMs. To further ensure consistency and reliability, we additionally employed a prompt template to instruct each model to cross-check the peer's output. Discrepant predictions were flagged for manual review and correction by domain experts. The detailed prompt designs for those stages are shown below.

\begin{tcolorbox}[
title=Task for Gemini\&GPT annotator, 
colback=white, 
colframe=black!70, 
arc=5pt,
fonttitle=\bfseries\ttfamily,
breakable,
pad at break=0pt,
bottomrule at break=0pt,
toprule at break=0pt
]
\scriptsize

You are an expert AI engine for materials science microscopy analysis. Your task is to analyze the provided micrograph and generate a structured JSON object. The analysis must be precise, and the final prompt must be concise and adhere to a strict format.

\vspace{1em}
\textbf{JSON Output Structure and Guidelines:}
\vspace{0.5em}

% --- 开始 JSON 结构 ---
\texttt{\{}
\begin{description}
    \item[\texttt{"subject"}:] Summary of the material type (e.g., \texttt{"Ceramic Powder"}, \texttt{"Metal Nanoparticles"}, \texttt{"Polymer Film"}).
    
    \item[\texttt{"microscopy\_type"}:] Identify the microscopy type: \texttt{"SEM"}, \texttt{"TEM"}, \texttt{"STEM"} or \texttt{"OM"}.
    
    \item[\texttt{"color\_profile"}:] Describe the foreground and background colors, including any false coloring or staining (e.g., \texttt{"Light gray particles on a dark background"}, \texttt{"Dark particles on a white background"}, \texttt{"Colorful stained specimen"}).
    
    \item[\texttt{"feature\_analysis"}:] \texttt{\{}
        \begin{description}
            \item[\texttt{"morphology"}:] Describe the fundamental shape of the individual particles/units (e.g., \texttt{"spherical"}, \texttt{"rod-shaped"}, \texttt{"angular polyhedral"}, \texttt{"fibrous"}, \texttt{"irregular polygonal"}, \texttt{"plate-like"}, \texttt{"nanowires"}).
            
            \item[\texttt{"particle\_density"}:] Estimate the particle count/density in the frame (e.g., \texttt{"single object"}, \texttt{"sparse"}, \texttt{"medium density"}, \texttt{"high density"}).
            
            \item[\texttt{"distribution"}:] Describe the spatial arrangement of particles (e.g., \texttt{"uniform"}, \texttt{"periodic"}, \texttt{"densely packed"}, \texttt{"isolated and scattered"}, \texttt{"agglomerated into clusters"}, \texttt{"interconnected network"}).
            
            \item[\texttt{"layering"}:] Describe the Z-axis arrangement. Choose ONE: \texttt{"Tiled"} (appears as a single flat layer) or \texttt{"Multilayer"} (particles are clearly stacked on top of each other).
            
            \item[\texttt{"surface\_texture"}:] Describe the texture on the surface of the individual particles (e.g., \texttt{"smooth"}, \texttt{"porous and web-like"}, \texttt{"crystalline facets"}, \texttt{"nanostructured"}).
            
            \item[\texttt{"pixel\_size\_profile"}:] Analyze the particle sizes as seen in the image pixels, focusing on uniformity. (e.g., \texttt{"Uniform particle sizes"}, \texttt{"Uniform in size with only minor variation"}, \texttt{"Wide range of particle sizes"}, \texttt{"Mostly uniform with occasional large outliers"}).
        \end{description}
    \hspace{1.5em}\texttt{\}}
\end{description}
\texttt{\}}
% --- 结束 JSON 结构 ---

\end{tcolorbox}

\begin{tcolorbox}[
title=Task for Gemini\&GPT cross-check, 
colback=white, 
colframe=black!70, 
arc=5pt,
fonttitle=\bfseries\ttfamily,
breakable,
pad at break=0pt,
bottomrule at break=0pt,
toprule at break=0pt
]
\scriptsize

You are an expert in materials science and microscopy image analysis. Given a microscopy image and a structured description of its contents, your task is to evaluate whether the description is accurate and consistent with the image.
\vspace{1em} % Adds a little vertical space

Please assess the consistency between the image and each of the following structured attributes:

\begin{itemize}
    \item \textbf{subject}: summary of the material type
    \item \textbf{microscopy\_type}: Identify the microscopy type
    \item \textbf{color\_profile}: Describe the foreground and background colors, including any false coloring or staining
    \item \textbf{morphology}: Describe the fundamental shape of the individual particles/units
    \item \textbf{particle\_density}: Estimate the particle count/density in the frame
    \item \textbf{distribution}: Describe the spatial arrangement of particles
    \item \textbf{layering}: Describe the Z-axis arrangement. Choose ONE: "Tiled" or "Multilayer"
    \item \textbf{surface\_texture}: Describe the texture on the surface of the individual particles
    \item \textbf{pixel\_size\_profile}: Analyze the particle sizes as seen in the image pixels, focusing on uniformity (e.g., "Uniform particle sizes", "Uniform in size with only minor variation", "Wide range of particle sizes", "Mostly uniform with occasional large outliers").
\end{itemize}

\vspace{1em}
Return a JSON object with the following format:
\vspace{0.5em}

% --- JSON Structure Start ---
\texttt{\{}
\begin{description}
    \item[\texttt{"overall\_consistency"}:] \texttt{"Yes"} or \texttt{"No"},
    
    \item[\texttt{"attribute\_consistency"}:] \texttt{\{}
        \begin{description}
            \item[\texttt{"subject"}:] \texttt{"Yes"} or \texttt{"No"},
            \item[\texttt{"microscopy\_type"}:] \texttt{"Yes"} or \texttt{"No"},
            \item[\texttt{"color\_profile"}:] \texttt{"Yes"} or \texttt{"No"},
            \item[\texttt{"morphology"}:] \texttt{"Yes"} or \texttt{"No"},
            \item[\texttt{"particle\_density"}:] \texttt{"Yes"} or \texttt{"No"},
            \item[\texttt{"distribution"}:] \texttt{"Yes"} or \texttt{"No"},
            \item[\texttt{"layering"}:] \texttt{"Yes"} or \texttt{"No"},
            \item[\texttt{"surface\_texture"}:] \texttt{"Yes"} or \texttt{"No"},
            \item[\texttt{"pixel\_size\_profile"}:] \texttt{"Yes"} or \texttt{"No"}
        \end{description}
    \hspace{1.5em}\texttt{\}},

    \item[\texttt{"comments"}:] \texttt{\{}
        \begin{description}
            \item[\texttt{"subject"}:] \texttt{"Optional brief explanation if inconsistent"},
            \item[\texttt{"microscopy\_type"}:] \texttt{"..."},
            \item[\texttt{...}]
        \end{description}
    \hspace{1.5em}\texttt{\}}
\end{description}
\texttt{\}}
% --- JSON Structure End ---
\end{tcolorbox}

\section{Segmentation Annotation}
\label{sec:annotation}
To enable reliable instance segmentation model training on EMs with densely packed instances, we developed a bespoke web‐based data engine that unifies dataset management, annotation, and quality assurance. Conventional annotation tools (e.g., LabelMe) struggle with our high‐density requirements—typical EMs with a resolution of 2,000 × 2,000 may contain over 2,000 discrete objects, leading to severe lag and application crashes. To address these limitations, we built an online platform specifically for EMs, featuring custom annotation widgets, a staged review workflow, and an iterative model–human feedback loop to concurrently expand both our dataset and model capabilities.

Our platform maintains fluid pan‐and‐zoom interactions and on‐demand rendering of regions of interest, even with tens of thousands of polygonal annotations in view. A key feature is the magnetic‐lasso tool with real‐time Sobel‐based edge snapping, which enables rapid and accurate tracing of complex, low‐contrast boundaries; annotators can dynamically adjust snapping sensitivity to accommodate variations in image contrast and material heterogeneity.

We enforce rigorous quality control through a multi‐phase review pipeline. Each image is first independently inspected by two peer annotators, who use inline comment threads to flag missing or misaligned masks. A senior materials‐science expert then conducts a final audit. Every comment is anchored to specific annotation vertices or instances, and all edits and review decisions are versioned, ensuring full traceability and the ability to revert changes if necessary.

To establish and progressively expand our dataset via model-assisted annotation, we adopted an iterative curriculum-learning framework. Initially, domain experts produced high-fidelity masks on a carefully curated, low-density subset of EMs. These gold-standard annotations trained a preliminary instance-segmentation network. In the subsequent iteration, we employed generative models to synthesize supplementary micrograph samples—thereby augmenting the training corpus and enhancing the network’s generalization capability—before applying the network to generate pseudo-labels on the remaining unannotated images. A larger cohort of trained volunteers then refined these pseudo-labels using our annotation tools, and all corrected masks were subjected to the established multi-stage quality-control workflow prior to incorporation into the final dataset.

\begin{figure}[ht]
  \centering
\includegraphics[width=0.8\textwidth]{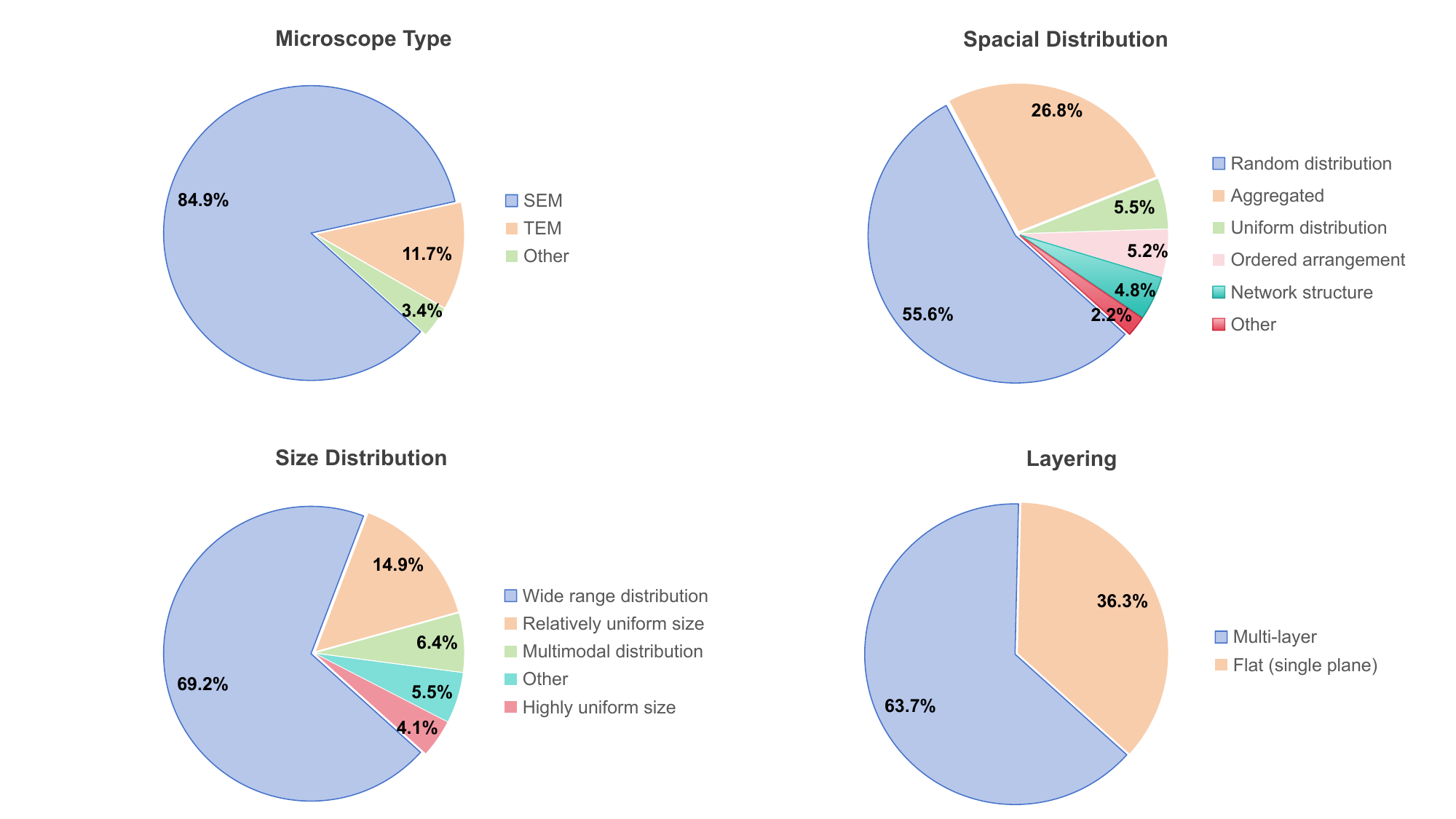}
\caption{Statistical distribution of four representative attributes from EM3M, illustrating its diversity: (a) Microscopy types, (b) Spatial distribution, (c) Size distribution, and (d) Layering in Z-axis.}
\label{fig:attribute_distribution}
\end{figure}

\section{Attribute Visualizations of EM3M}
\label{sec:attri_vis}
In this section, we present additional attribute distributions that further characterize the EM3M, as shown in Fig.~\ref{fig:attribute_distribution}. Specifically, we visualize four dimensions: microscopy type, spatial distribution, microstructure size distribution, and layering. Similar to the main text, we applied a coarse-grained categorical scheme to accommodate the substantial diversity in the data, aggregating samples into interpretable high-level groups.

As observed in Fig.~\ref{fig:attribute_distribution}(a), the dataset is dominated by SEM (scanning electron microscopy) images, which constitute over 90\% of the samples, with limited but non-negligible coverage of TEM (transmission electron microscopy) and other modalities. Spatial distribution patterns (Fig.~\ref{fig:attribute_distribution}(b)) reveal substantial heterogeneity, with random and aggregated arrangements being most prevalent, though uniform and ordered structures are also represented. In terms of particle size variation (Fig.~\ref{fig:attribute_distribution}(c)), a wide range distribution is most common, while subsets with relatively uniform or multimodal sizes offer additional diversity. The layering property (Fig.~\ref{fig:attribute_distribution}(d)) shows a meaningful split between multi-layered and flat (single-plane) micrographs, further contributing to the structural variability of the EM3M.

\begin{figure}[htbp]
  \centering
  \begin{subfigure}{0.3\linewidth}
    \includegraphics[width=\linewidth]{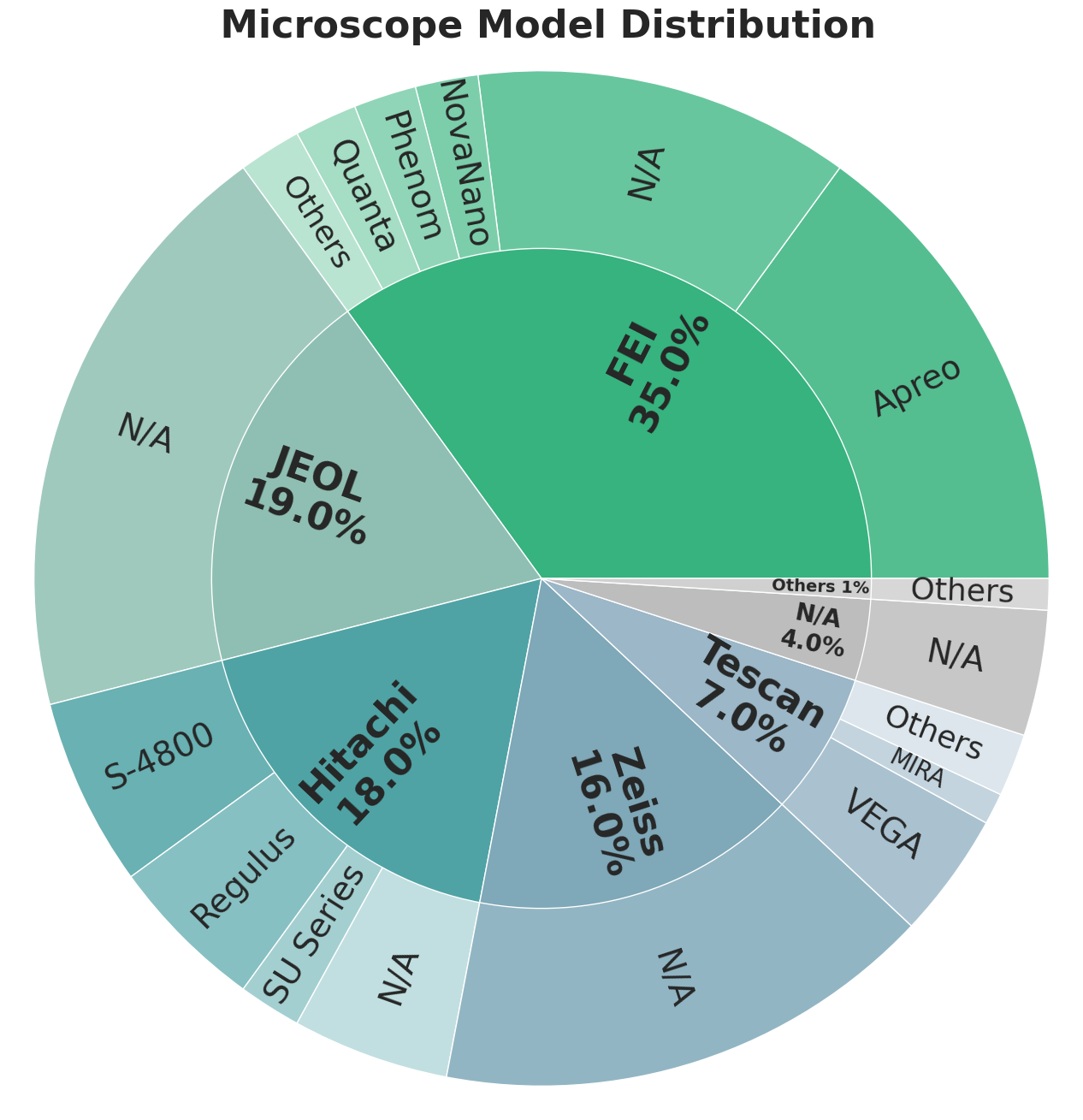}
    \caption{}
    \label{fig:micro_brands}
  \end{subfigure}
  \hfill
  \begin{subfigure}{0.33\linewidth}
    \includegraphics[width=\linewidth]{{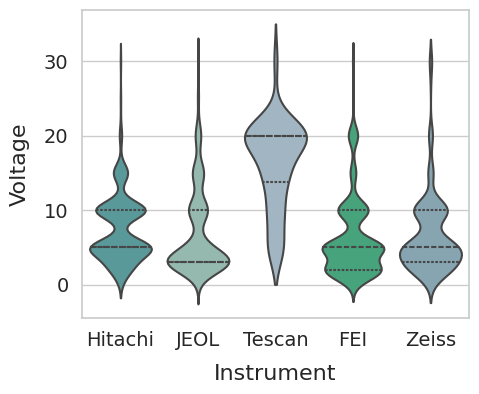}}
    \caption{}
    \label{fig:voltage_plot}
  \end{subfigure}
  \hfill
  \begin{subfigure}{0.33\linewidth}
    \includegraphics[width=\linewidth]{{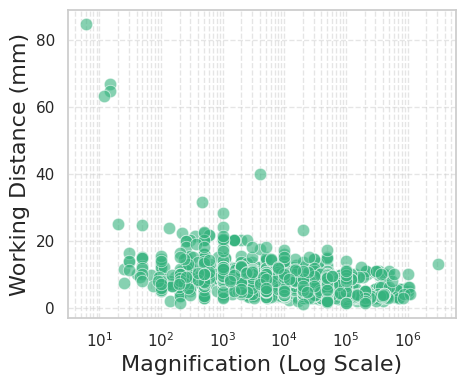}}
    \caption{}
    \label{fig:WD_Mag}
  \end{subfigure}

  \caption{Statistical overview of the metadata extracted from the microscopy dataset. (a) Distribution of microscope manufacturers and specific models. (b) Distribution of operating voltages across different instrument brands. (c) Relationship between working distance and magnification (log scale).}
  \label{fig:metadata_vis}
\end{figure}

On the other hand, Our dataset features exceptional diversity, aggregating electron microscopy images from more than 60 research institutions and covering a broad spectrum of instrument models and imaging configurations. By systematically extracting and parsing the metadata from the image data bars, we quantified the technical variations across the dataset. As illustrated in Fig.~\ref{fig:metadata_vis}, the statistical results summarize the prevalent hardware distributions (e.g., manufacturers and models) and key operational parameters (e.g., accelerating voltage, working distance, and magnification).

\begin{figure}[htbp]
  \centering
\includegraphics[width=0.7\textwidth]{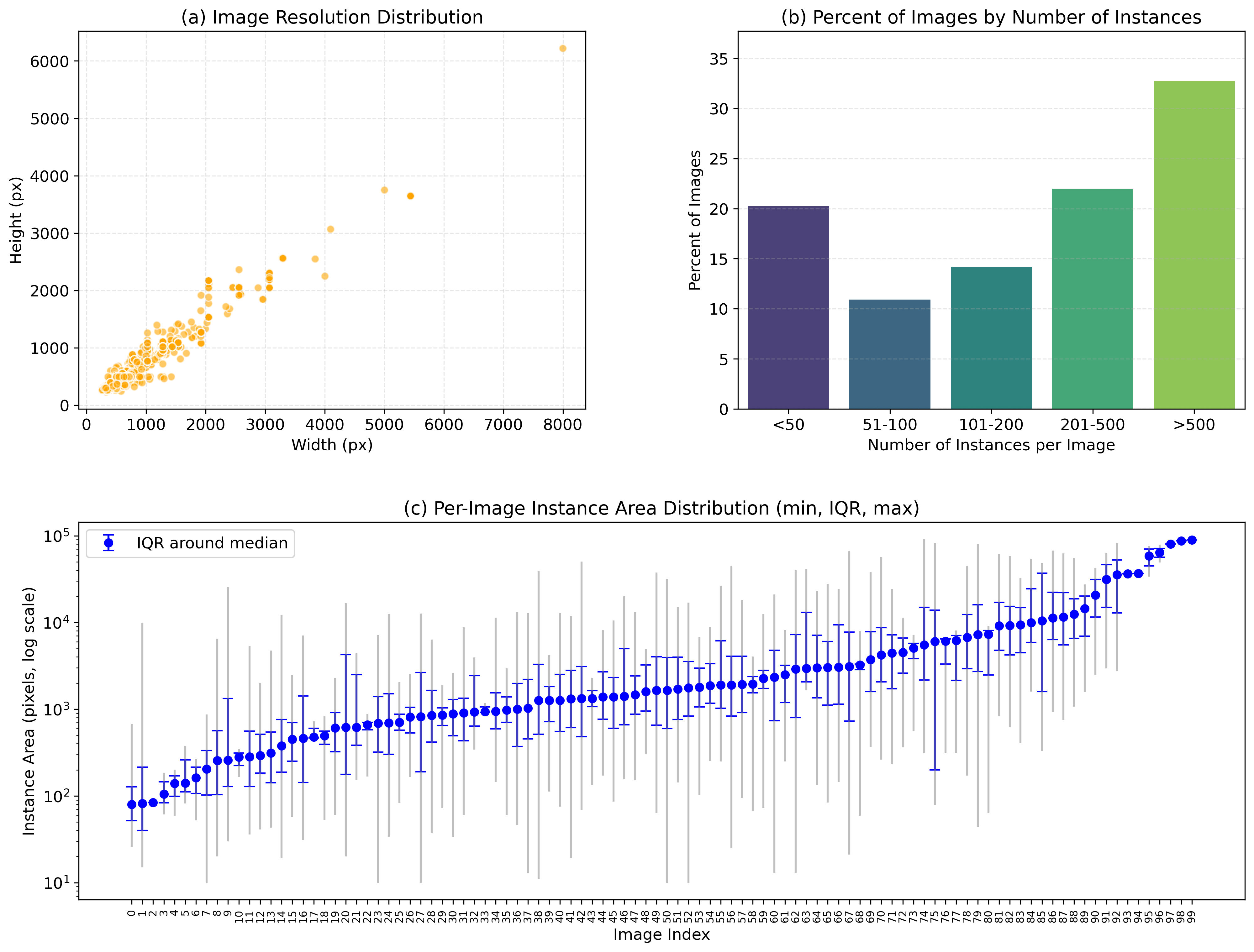} % Reduce the figure size so that it is slightly narrower than the column. Don't use precise values for figure width.This setup will avoid overfull boxes.
\caption{Statistical distribution of EM3M. (a) Distribution of image resolution. (b) Percentage of images by number of instances. (c) Per-image instance area range (min, IQR, max, log scale).}
\label{fig:instance_distribution}
\end{figure}

\section{Statistics Distribution of EM3M}
\label{sec:distribution}
Fig.~\ref{fig:instance_distribution}(a) shows the distribution of image resolutions, revealing a wide range of sizes—from below 500 pixels in width and height to over 8000 × 6000 pixels—reflecting significant variability in source data quality and acquisition conditions. Fig.~\ref{fig:instance_distribution}(b) summarizes the proportion of images by instance count, indicating that over 30\% of images contain more than 500 annotated instances, highlighting the dataset’s propensity toward highly dense scenes. Finally, Fig.~\ref{fig:instance_distribution}(c) visualizes the distribution of instance areas across 100 randomly selected images. For each image (sorted by median instance size), the minimum, maximum, and interquartile range (IQR) are shown on a log scale. The broad spread of instance areas within and across images further underscores the substantial intra- and inter-image scale variation, reinforcing the challenges posed to segmentation models.

%%%%%%%%%%%%%%%%%%%%%%%%%%%%%%%%%%%%%%%%%%%%%%%%%%%%%

\section{Implementation Details}
\label{sec:implementation}
\subsection{Instance Segmentation}
\noindent
\textbf{Common Settings.}\quad To ensure rigor and reproducibility, we use publicly available official codebases. Given the diverse architectures of the evaluated methods, we adopt the most appropriate and well-supported implementation for each rather than unifying them under a single framework. The corresponding codebases are listed in Tab.~\ref{tab:codebases}. All models are trained on our datasets until full convergence. Specifically, models are trained for 180,000 iterations on the full EM3M dataset and for 60,000 iterations on the sparse subset, using a batch size of 8 distributed across 4 NVIDIA 4090D GPUs. Data augmentation includes random horizontal and vertical flipping, random cropping with varying scales, and resizing to 1024$\times$1024. For each method, we start from its default training configuration and apply minimal but targeted modifications to adapt it to our dataset.

\begin{table}[htbp]
\centering
\scriptsize
\caption{Codebases used for implementing the benchmarked methods.}
\label{tab:codebases}
\newcolumntype{L}{>{\raggedright\arraybackslash}X}
\begin{tabularx}{\linewidth}{@{}lL@{}}
\toprule
\textbf{Codebase} & \textbf{Implemented Methods} \\ \midrule
MMDetection~\cite{chen2019mmdetection} & Mask R-CNN, Cascade R-CNN, HTC, YOLACT, SOLOv2, Mask2Former \\ \addlinespace
CellViT Codebase~\cite{horst2024cellvit} & StarDist, CPPNet, HoverNet, CellViT \\ \addlinespace
Detectron2~\cite{wu2019detectron2} & MaskDINO, FastInst, UniAIMS \\ \addlinespace
Ultralytics YOLO~\cite{Jocher_Ultralytics_YOLO_2023} & YOLO11-seg \\ \addlinespace
Cellpose Codebase~\cite{pachitariu2025cellpose} & Cellpose \\ \bottomrule
\end{tabularx}
\end{table}

\noindent
\textbf{Detection-based and Query-based Methods.}\quad
For methods implemented within the MMDetection framework (Mask R-CNN, Cascade R-CNN, HTC, YOLACT, SOLOv2, and Mask2Former), we follow a consistent training protocol. CNN-based models are trained using the SGD optimizer with a momentum of 0.9 and a weight decay of 0.0001. The schedule includes a 500-iteration linear warm-up phase where the learning rate increases from 0 to 0.005, followed by decay by a factor of 0.1 at 24k, 48k, and 54k iterations. Mask2Former is trained with AdamW (initial LR of 1e-4, weight decay 0.05), using a 1.5k-iteration warm-up and cosine annealing for the remainder. YOLO11-seg, MaskDINO and FastInst are trained using their official repositories with default hyperparameters.

\noindent
\textbf{Bottom-up Methods.}\quad
Following recent works like CellViT~\cite{horst2024cellvit}, all bottom-up models are equipped with a SAM-pretrained ViT backbone to leverage strong feature representations. For StarDist, CPPNet, HoverNet, and CellViT, we use the comprehensive CellViT codebase as a unified framework for their comparative study. Models within this group are trained using the AdamW optimizer with an initial learning rate of 1e-4, incorporating a 2000-iteration linear warm-up phase followed by a cosine annealing schedule for the remainder of training. For Cellpose, its official public codebase is employed using its default parameter configurations. Given that the Cellpose implementation does not natively support multi-GPU training, its experiments is conducted on a single GPU. To accommodate this, a batch size of 8 and an input image resolution of 256×256 pixels is adopted for the benchmark, ensuring fidelity to its standard usage while addressing hardware constraints. Our proposed baseline, UniAIMS, builds directly upon the flow-based paradigm of Cellpose but is engineered to overcome these limitations. We address the scalability issues with multi-GPU distributed training support and enhance its robustness by replacing the original size-model with LSJ augmentations, making it highly effective for the significant scale variation present in our dataset.

\subsection{Justification for the Sparse Subset Evaluation}
\label{sec:Justification}
The full EM3M dataset was first partitioned into a training set (4,128 images) and a test set (963 images). From these splits, we curated a sparse subset by selecting all images containing fewer than 100 annotated instances. This process resulted in a sparse training set of 1,210 images and a sparse test set of 371 images, which were used exclusively for benchmarking detection-based and query-based methods. 

Our primary motivation was to establish a fair benchmark using the default configurations provided by major codebases like MMDetection. Most canonical instance segmentation models are configured by default with a maximum detection limit of 100 instances per image. By adhering to these standard settings, we avoid dataset-specific hyperparameter tuning and instead report the practical, out-of-the-box performance of these established methods.

Furthermore, it is a well-documented consensus that the architectural designs of top-down and query-based models are not well-suited for extremely dense scenes. Core mechanisms, such as NMS in detection-based models and the fixed number of object queries in transformer-based models, become severe performance bottlenecks when handling hundreds or thousands of highly overlapping objects. Our evaluation, therefore, focuses on the sparse subset where these methods can operate within their intended design limits, allowing for a meaningful and direct comparison of their segmentation quality.

\begin{figure}[t]
\centering
\includegraphics[width=0.85\columnwidth]{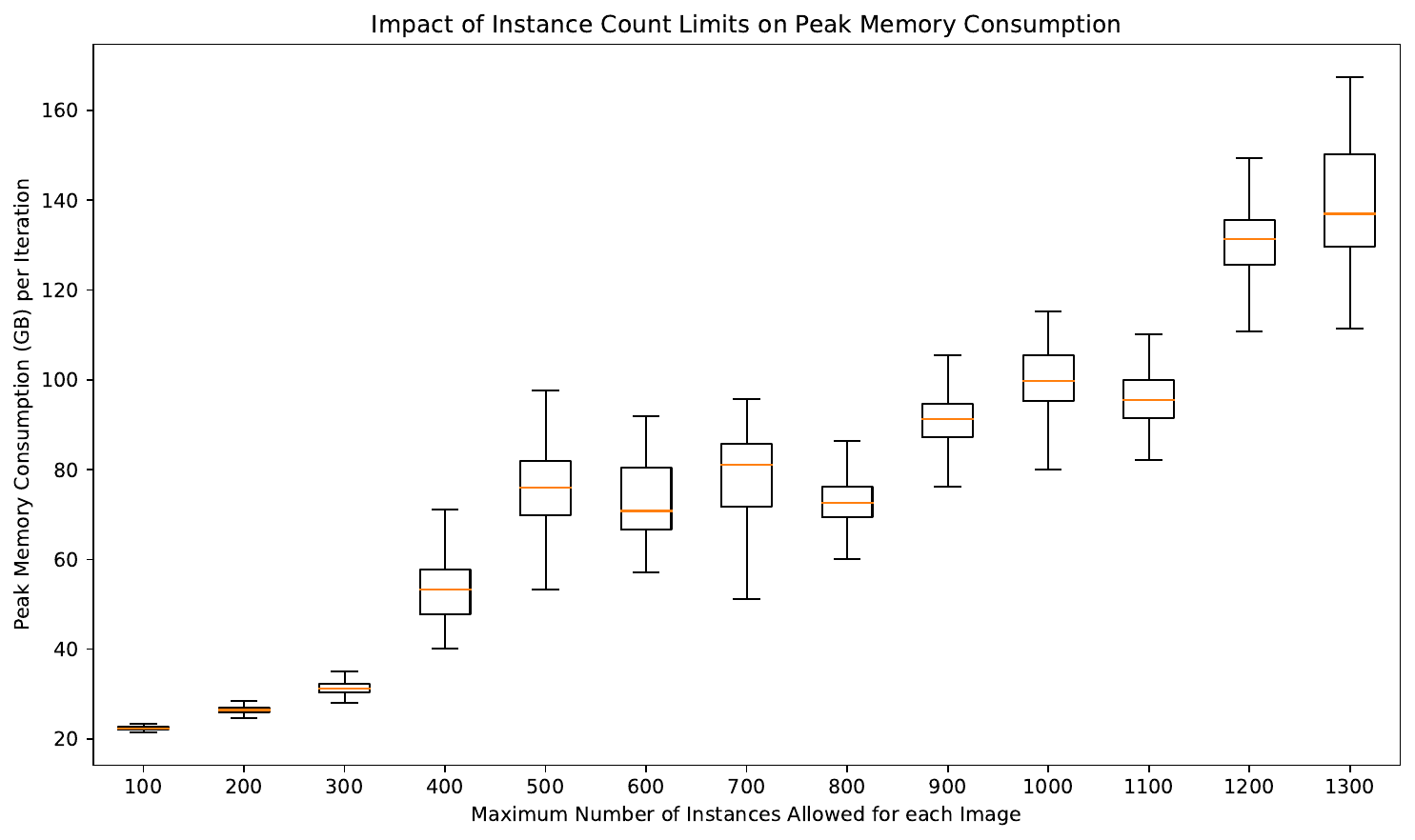} % Reduce the figure size so that it is slightly narrower than the column. Don't use precise values for figure width.This setup will avoid overfull boxes.
\caption{Impact of instance count limits on peak memory consumption.}
\label{fig:peak_memory}
\end{figure}

To provide a concrete illustration of the practical hardware constraints, we analyzed the peak system memory (RAM) consumption of a Mask R-CNN model within the MMDetection framework. After an initial warm-up period for memory usage to stabilize, we recorded the peak RAM usage over 1,000 training iterations while systematically increasing the instance count limit. As shown in Fig.~\ref{fig:peak_memory}, the memory overhead during the data loading pipeline increases dramatically with the number of instances. Due to the random nature of batch sampling, a single training batch can occasionally be composed of multiple high-density images. This creates a worst-case scenario that can easily trigger an Out-of-Memory (OOM) error, making stable training on the full dataset intractable. While this plot focuses on system RAM, a similar, albeit less pronounced, trend was also observed for GPU memory (VRAM) consumption. This experiment empirically confirms that these architectures are mechanically ill-suited for the scale of our full dataset, further justifying our decision to benchmark them on the sparse subset.

\section{Prediction Results}
\label{sec:prediction}
\begin{figure*}[htbp]
\centering
\includegraphics[width=0.95\textwidth]{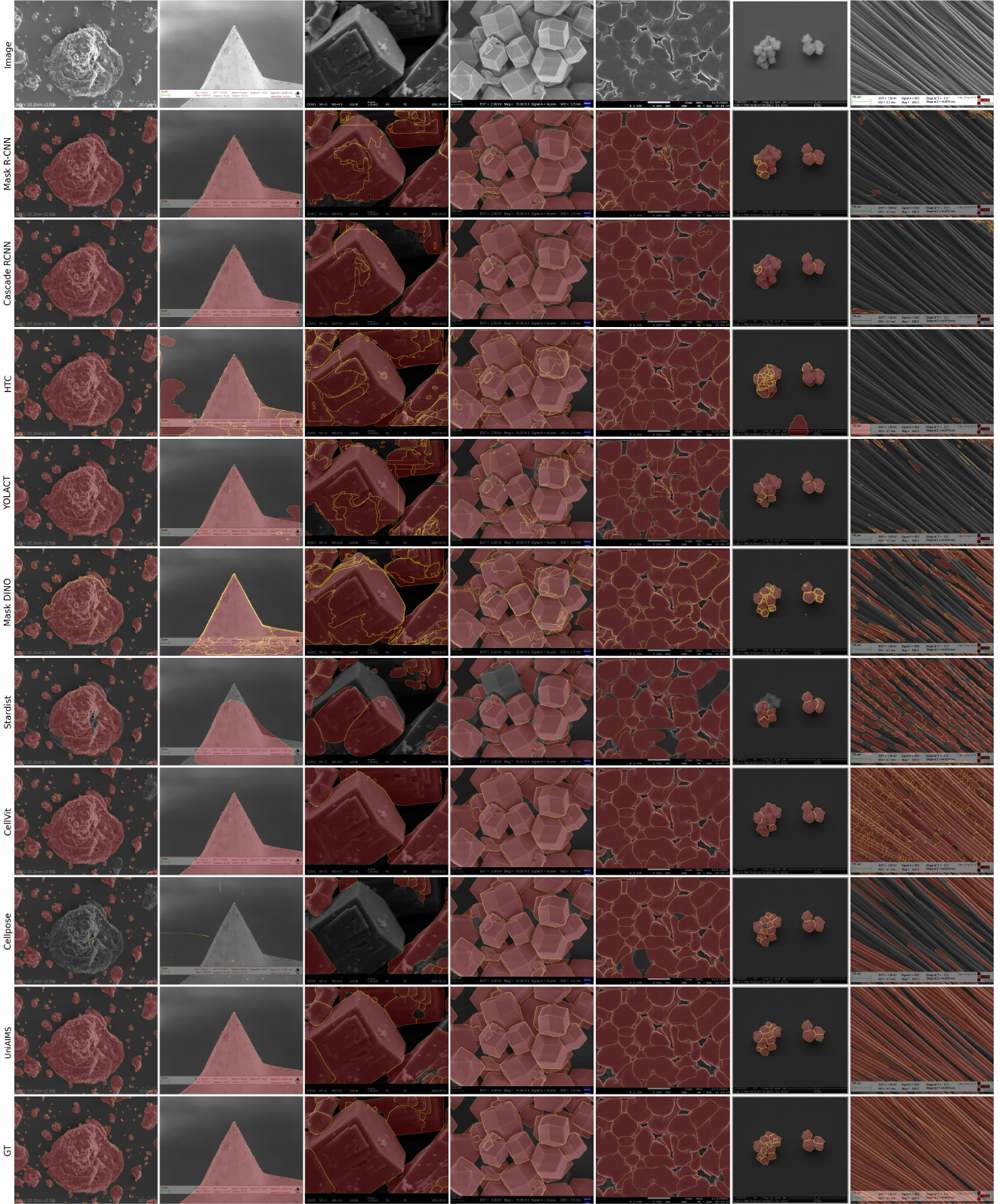} % Reduce the figure size so that it is slightly narrower than the column. Don't use precise values for figure width.This setup will avoid overfull boxes.
\caption{Visual results of several instance segmentation methods on sparse subset of EM3M.}
\label{fig:main_results}
\end{figure*}

\noindent
\textbf{Qualitative Analysis.}\quad
Fig.~\ref{fig:main_results} presents a comprehensive qualitative comparison of representative instance segmentation frameworks evaluated on a sparse subset of the EM3M dataset. The seven micrograph columns span a diverse range of morphological contexts, from irregular agglomerates to sharp-edged crystalline structures and fine fibrous arrays, each chosen to probe distinct segmentation challenges such as multi-scale feature extraction, boundary preservation, and instance separation under occlusion.

\noindent
\textbf{Detector-based and Query-based Methods.}\quad
These paradigms, such as Mask R-CNN, Cascade R-CNN, and Mask DINO, exhibit consistent weaknesses in dense or geometrically complex scenarios. They often produce incoherent or fragmented masks and struggle to effectively separate adjacent instances in cluttered regions (Columns 3–6). Furthermore, as seen in Column 2, these methods fail to preserve the sharp, linear facets of the pyramidal crystal, instead generating overly smoothed contours that compromise morphological fidelity.

\noindent
\textbf{Bottom-up Methods.}\quad
While bottom-up approaches show better adaptability to dense microstructures, each still exhibits characteristic shortcomings. StarDist produces coarse, imprecise contours around irregular particles, indicating low boundary accuracy. CellViT fails to effectively separate touching or partially overlapping grains, leading to under-segmentation in Columns 3 and 6. Conversely, Cellpose successfully resolves many dense instances but often omits large or low-contrast particles, as visible in Column 1-3. Collectively, these results highlight that bottom-up pipelines, though inherently better suited for dense textures, still struggle with geometric precision and object completeness in heterogeneous EM scenes. Moreover, the final fibrous sample exposes a shared failure mode across nearly all baselines: the inability to delineate elongated, repetitive structures against a textured background. Here, every prior method collapses to near-complete failure, producing either blank masks or heavy mis-segmentation. 

\noindent
\textbf{UniAIMS.}\quad
In contrast, our proposed UniAIMS demonstrates remarkable robustness and precision across all evaluated scenarios. It is the only framework that accurately preserves the sharp geometric features of the crystal (Column 2) while simultaneously resolving individual instances within the most densely packed regions (Columns 3–5). Most notably, it uniquely overcomes the universal failure mode observed in the fibrous texture case (Column 7), highlighting its superior capability to delineate complex structures against challenging backgrounds.

\end{document}